\pdfoutput=1
\documentclass[letterpaper, 10 pt, conference]{ieeeconf}  

\IEEEoverridecommandlockouts                              

\newtheorem{theorem}{Theorem}

\usepackage[utf8]{inputenc} 
\usepackage{amsmath,amssymb,amsfonts}
\usepackage{mathtools}
\usepackage{graphicx}
\usepackage{textcomp}
\usepackage{xcolor}
\usepackage{booktabs} 
\usepackage{multirow}
\usepackage{caption}
\usepackage{subfig}
\usepackage{float} 
\usepackage{microtype}
\usepackage{nicefrac}       
\usepackage{cite}
\usepackage[algo2e,noend]{algorithm2e}
\usepackage{algorithm}  
\usepackage{algpseudocode}  
\usepackage[noend]{algcompatible}
\usepackage[hyphens]{url}  
\DeclareMathOperator*{\argmax}{arg\,max}

\newcommand{\bspi}{{\boldsymbol{\pi}}}
\newcommand{\bsmu}{{\boldsymbol{\mu}}}
\newcommand{\cX}{{\mathcal{X}}}
\newcommand{\cA}{{\mathcal{A}}}
\definecolor{softblue}{rgb}{0.2, 0.2, 0.6} 
\newcommand{\defi}[1]{{\bf #1}}

\def\BibTeX{{\rm B\kern-.05em{\sc i\kern-.025em b}\kern-.08em
    T\kern-.1667em\lower.7ex\hbox{E}\kern-.125emX}}

\makeatletter 
\let\NAT@parse\undefined
\makeatother
\usepackage[colorlinks = true,
            linkcolor = blue,
            urlcolor  = blue,
            citecolor = blue,
            anchorcolor = blue]{hyperref} 
            
\title{\LARGE \bf
Population-aware Online Mirror Descent for Mean-Field Games with Common Noise by Deep Reinforcement Learning}

\author{Zida Wu$^{1}$, Mathieu Lauriere$^{2}$, Matthieu Geist$^{3}$, Olivier Pietquin$^{4}$, Ankur Mehta$^{1}$%
\thanks{$^{1}$UCLA {\tt\small \{zdwu,mehtank\}@ucla.edu}}%
\thanks{$^{2}$New York University Shanghai. {\tt\small mathieu.lauriere@nyu.edu}}%
\thanks{$^{3}$Cohere. {\tt\small matthieu.geist@univ-lorraine.fr}}%
\thanks{$^{4}$University of Lille. {\tt\small olivier.pietquin@univ-lille.fr}}%
}

\begin{document}

\maketitle
\thispagestyle{empty}
\pagestyle{empty}

\begin{abstract}
Mean Field Games (MFGs) offer a powerful framework for studying large-scale multi-agent systems. Yet, learning Nash equilibria in MFGs remains a challenging problem, particularly when the initial distribution is unknown or when the population is subject to common noise. In this paper, we introduce an efficient deep reinforcement learning (DRL) algorithm designed to achieve population-dependent Nash equilibria without relying on averaging or historical sampling, inspired by Munchausen RL and Online Mirror Descent. The resulting policy is adaptable to various initial distributions and sources of common noise. Through numerical experiments on seven canonical examples, we demonstrate that our algorithm exhibits superior convergence properties compared to state-of-the-art algorithms, particularly a DRL version of Fictitious Play for population-dependent policies. The performance in the presence of common noise underscores the robustness and adaptability of our approach.
\end{abstract}

\section{Introduction}
\label{sec:introduction}
Multi-agent systems (MAS) \cite{dorri2018multi} are common in real-life scenarios involving many players, such as flocking \cite{olfati2006flocking, cucker2007emergent}, traffic flow \cite{burmeister1997application}, and swarm robotics \cite{chung2018survey}. While many MAS models are generally pure dynamical systems, other models incorporate rationality through game theory. As the number of players grows, computational efficiency becomes a major challenge \cite{lowe2017multi,rashid2020monotonic}. However, under symmetry and homogeneity assumptions, mean field approximations offer an effective way to model population behaviors and learn decentralized policies, avoiding the curse of dimensionality.

Mean field games (MFGs) \cite{lasry2007mean,huang2007large,carmona2018probabilistic,bensoussan2013mean} provide a framework for large-population games where agents interact through the population distribution. As the number of agents grows, individual influence diminishes, reducing interactions to those between a representative agent and the population. The solution concept in MFGs is the Nash equilibrium, where no player has an incentive to deviate unilaterally. Learning methods for MFGs often rely on fixed-point iterations, which update agent policies and mean-field terms iteratively. However, convergence of these methods requires strict contraction conditions \cite{li2019efficient, guo2019learning}, which often fail in practice \cite{cui2021approximately, anahtarci2023q}.

To address these limitations, smoothing-based approaches such as Fictitious Play (FP) have been introduced. Originally developed for finite-player games \cite{brown1951iterative,berger2007brown}, FP has been extended to MFGs \cite{cardaliaguet2017learning,hadikhanloo2019finite,perrin2020fictitious} (see Algo.~\ref{algorithm2}). FP averages historical distributions, and can be proved to converge under a structural assumption (Lasry-Lions monotonicity) instead of a strict contraction property. However, FP faces computational challenges. Averaging policies is difficult with non-linear approximators like deep neural networks (DNNs) \cite{lauriere2022scalable}, and computing best responses for each iteration is costly. Additionally, uniform averaging over all past iterations slows the update rate as iterations increase.

Online Mirror Descent (OMD) addresses these issues by aggregating past Q-functions instead of policies \cite{hadikhanloo2017learning,hadikhanloo2018learning,perolat2021scaling} (see Algo.~\ref{algorithm3}). Unlike FP, OMD only requires policy evaluation, not optimization, and maintains a constant update rate. Although summing explicitly Q-functions is costly with DNNs, \cite{lauriere2022scalable} proposed a deep RL adaptation using the Munchausen trick \cite{vieillard2020munchausen} which provides implicit regularization.

Most of the MFG literature focuses on computing Nash equilibrium for a single mean field (i.e., one distribution sequence); see~\cite{lauriere2022learning} for a survey. In such cases, decentralized policies that depend only on individual states and are myopic to the population distribution are sufficient. However, this assumption restricts MFG theory’s applicability. The \emph{Master equation} has been introduced to study the individual value function as a function of the population distribution~\cite{cardaliaguet2019master}. \cite{Perrin2022General} \cite{wu2024population} proposed to compute \emph{master policies}, enabling Nash equilibrium computation from any initial distribution. However, FP-type algorithms face intrinsic limitations, such as costly best-response computations and decaying learning rates.

{\bf Main challenges. } The challenges are threefold: policies should be \emph{population-dependent}, the algorithm should handle \emph{unknown initial distribution and common noise}, and the problem is set in \emph{finite horizon} with non-stationary policies.

{\bf Main contributions. }  We propose, \defi{Master OMD (M-OMD)}, a \emph{population-dependent DRL algorithm} which is able to handle unknown initial distributions. We extend the algorithm to handle \emph{common noise} impacting the whole population distribution. Last, we demonstrate through extensive \emph{numerical experiments} that M-OMD has better generalization property than state-of-the-art baselines.



\section{Problem Definition}
\label{sec:background}
In this paper, we consider a multi-agent decision-making problem using the framework of discrete-time MFGs, which relies on the classical notion of Markov decision processes (MDPs). {\bf Notations:} For any finite set $E$, we denote by $\Delta_E$ the simplex of probability distributions on a $E$. For any integer $N$, we denote $[N] = \{0,1,\dots,N\}$. Functions of time are seen as sequences and denoted by bold letters.

\paragraph{MDP for a representative agent.} 
We first present the setting without common noise and extend it to common noise in Section~\ref{sec:common-noise}. The state space $\mathcal{X}$ and action space $\mathcal{A}$ are finite, with a time horizon $N_T$. At each step $n \in [N_T]$, the mean field term (population's state distribution) is $\mu_n \in \Delta_{\mathcal{X}}$. When an agent in state $x_n$ takes action $a_n$, the next state is determined by the transition probability $p_n(\cdot\mid x_n, a_n, \mu_n)$, and the reward is $r_n(x_n, a_n, \mu_n)$. We focus on population-independent transitions, though our algorithm can handle population-dependent cases. The mean field sequence is denoted as $\bsmu = (\mu_n)_{n \in [N_T]}$. Given this sequence, the agent aims to maximize cumulative rewards up to $N_T$. Unlike most of the MFG literature, the initial distribution $\mu_0$ will be \emph{variable}. This motivates the following class of policies.


\paragraph{Classes of policies.} 
We use \emph{population-dependent} policies, also called \defi{master policies} \cite{Perrin2022General}: $\boldsymbol{\pi} = (\pi_n)_{n \in [N_T]}$ with $\pi_n: \mathcal{X} \times \Delta_{\mathcal{X}} \to \Delta_{\mathcal{A}}$. If at time $n$ the player is in state $x_n$ and the mean field is $\mu_n$, the player picks an action $a_n \sim \pi_n(\cdot|x,\mu)$ where $\pi_n(\cdot|x,\mu)$ is equivalent as $\pi_n(\cdot|x_n,\mu_n)$.

\paragraph{Best Response.} 
From the perspective of a single agent, if the mean field sequence $\bsmu = (\mu_n)_{n\in[N_T]}$ is given, the total reward function to maximize is defined as:
\begin{equation*}
\textstyle
    J(\bspi ; \bsmu)=\mathbb{E}\Big[\sum_{n \in [N_T]} r_n\left(x_n, a_n, {\mu}_n\right)\Big],
\end{equation*}
subject to the dynamics $ x_0 \sim \mu_0$, $x_{n+1} \sim p_n\left(\cdot \mid x_n, a_n, \mu_n\right)$, $a_n \sim \pi_n\left(\cdot \mid x_n, \mu_n\right)$, for $n \in[N_T-1]$. A policy  $\bspi$ is a \defi{best response} against $\bsmu$ if it maximizes  $J$, i.e.,
\(
    \bspi \in BR(\bsmu):=\argmax_{\bspi} J(\bspi ; \bsmu). 
\)

\paragraph{Mean field. } We denote by $\bsmu^{\mu_0,\bspi} = MF_{\mu_{0}}(\bspi)$ the \defi{mean field generated by} policy $\bspi$ when starting from $\mu_0$. It is interpreted as the sequence of population distributions obtained when the population starts with $\mu_0$ and every player uses $\bspi$. It is defined as: for $n \in [N_T-1]$, 
\looseness=-1
\begin{equation}
\label{eq:MF-induced-pi}
    \textstyle
    \mu_{n+1}^{\mu_0,\bspi}(x') = \sum_{x,a} \mu_n^{\mu_0,\bspi}(x)\pi_n(a|x,\mu_n^{\mu_0,\bspi}) p_n\left(x' \mid x, a, \mu_n^{\mu_0,\bspi}\right).
\end{equation}

\paragraph{Nash equilibrium and master policies. } For a given initial $\mu_0$,
a pair $(\bspi, \bsmu)$ is a \defi{(mean-field) Nash equilibrium (MFNE)} if $\bspi$  is a best response to $\bsmu$, and $\bsmu$ is generated by $\bspi$ starting from $\mu_0$. Mathematically, $\bspi \in BR(\bsmu)$ and $\bsmu = \bsmu^{\mu_0,\bspi} = MF_{\mu_{0}}(\bspi)$. $\bspi$ is an equilibrium policy for a given initial $\mu_0$ if $\bspi \in BR(\bsmu^{\mu_0,\bspi})$, which means $\bspi$ is a fixed point of $BR \circ MF_{\mu_0}$. 
A population-dependent policy is called a {\bf master policy} if it is an equilibrium policy for any initial distribution $\mu_0$, i.e., $\bspi \in \cap_{\mu_0 \in \Delta_\mathcal{X}} BR(\bsmu^{\mu_0,\bspi})$. If the players use $\bspi$, then for any $\mu_0$, the population is in an MFNE.

\paragraph{Exploitability. }
In general, we cannot measure the distance between a given policy and the equilibrium one since it is unknown. Exploitability measures how far a policy is from being a Nash equilibrium~\cite{lockhart2019computing,perrin2020fictitious} by quantifying the maximum benefit a player can achieve by deviating from the policy used by the population: 
\begin{equation}
\label{eq:exploitabiliy}
    \textstyle \mathcal{E}^{\mu_0}(\bspi) = 
   \mathbb{E}_{\mu_0} \left[  \sup_{\bspi^{\prime}} J\left(\bspi^{\prime} ; \bsmu^{\mu_0,\bspi}\right)-J_{}\left(\bspi ; \bsmu^{\mu_0,\bspi}\right) \right].
\end{equation}
A policy $\bspi$ is a Nash equilibrium policy for $\mu_0$ if and only if $\mathcal{E}^{\mu_0}(\bspi) = 0$. 

\paragraph{Challenges and proposed approach. } To learn a master policy, there are three main challenges. First, the policy \emph{takes as input a population distribution}, which is a possibly high-dimensional vector. We will used a deep neural network to approximate it. Second, the policy should be an \emph{optimal policy for a rather complex MDP}. For this we will use model-free reinforcement learning method, which is more scalable than exact dynamic programming methods. Last, the policy needs to be able to \emph{perform well on any population distribution}. To achieve this, we will employ a training set composed of various initial distributions. Furthermore, to deal with the challenge of \emph{common noise from the environment}, the goal is to identify the minimal and necessary information required to reach the Nash equilibrium. We prove that our algorithm can naturally extend to the case in which incorporating the common noise as one input component is sufficient for the policy to converge to the Nash equilibrium.

\section{Algorithm}
\label{sec:algo}
In this section, we present our algorithm. 

\paragraph{Online Mirror Descent. }
Our approach builds upon the Online Mirror Descent (OMD) algorithm which was introduced for MFGs with population-independent policies in~\cite{perolat2021scaling}. It is inspired by the Mirror Descent MPI algorithm~\cite{vieillard2020leverage} in the single-agent case. At each iteration $k$, the policy $\boldsymbol\pi^{k-1}$ from the previous iteration is known. We first compute the mean field $\bsmu^{k-1} = \bsmu^{\bspi^{k-1}}$. Second, we evaluate $\boldsymbol\pi^{k-1}$ by computing its Q-function $Q^k = Q^{\bspi^{k-1}}$. Then, we compute the cumulative Q-function $\bar{Q}^k_n(x,a) = \sum_{i=0}^k Q^i_n(x,a)$. Finally, the new policy is computed as: $\pi^k_n(\cdot|x) = \operatorname{softmax}\left(\frac{1}{\tau} \bar{Q}^k(x,\cdot)\right)$. Since the policy depends only on the individual state and $\cX$ is finite, a tabular implementation has been used in~\cite{perolat2021scaling}. However, here our goal is to learn a master policy, which is \emph{population-dependent}. Since $\mu \in\Delta_{\mathcal{X}}$ takes a continuum of values, it is not possible to represent exactly $Q^k$. 
For this reason, we will resort to function approximation and use DNNs to approximate Q-functions. Classically, learning the Q-function associated with a policy can be done using Monte Carlo samples. Nevertheless, computing the sum of neural network Q-functions would be challenging because DNNs are non-linear approximators. To tackle this challenge, as explained below, we will use a similar idea as Munchausen trick, introduced in~\cite{vieillard2020munchausen} and adapted to the MFG setting in~\cite{lauriere2022scalable}. Although the latter reference also uses an OMD-type algorithm, it does not handle population-dependent policies and Q-functions, which is a major difficulty addressed here with suitable use of an initial distributions training set and of the replay buffer. 

\paragraph{Q-function update. } 
To compute $\bar{Q}$, a naive implementation would consist in keeping copies of past DNNs $(Q^i)_{i \in [k]}$, evaluating them and summing the outputs but this would be extremely inefficient both in terms of memory and computation. Instead, we define a regularized Q-function and establish Thm~\ref{thm:equivalence} similar to use the Munchausen trick~\cite{vieillard2020munchausen,lauriere2022scalable}. However, we derive the conclusion differently, which allows the target policy to be updated periodically during learning, similar to DQN, rather than being fixed as the policy learned from the last iteration. This change enhances the stability during training. It relies on computing a single Q-function that mimics the sum $\sum_{i=0}^{k-1} Q^i$ by using implicit regularization thanks to a Kullback-Leibler (KL) divergence between the new policy and the previous one. We derive, in our population-dependent context, the equivalence between regularized Q and the summation of historical Q values. 

\begin{theorem} 
\label{thm:equivalence}
Let $\tau>0$. Denote by $\bspi^{k-1}$ the softmax policy learned in iteration $k-1$, i.e., $\pi^{k-1}_n(\cdot|x,\mu) = \operatorname{softmax}\big(\frac{1}{\tau} \sum_{i=0}^{k-1} Q^i_n(x,\mu,\cdot)\big)$, and by $Q^k = Q^{\bspi^{k-1}}$  the state-action value function in iteration $k$. Let $\widetilde{Q}^k=Q^k+\tau \ln \bspi^{k-1}: \mathbb{N} \times \mathcal{X} \times \Delta_{\mathcal{X}} \times \mathcal{A} \to \mathbb{R}$. If $\mu^k$ is generated by $\bspi^{k-1}$ in Algo.~\ref{algo: algorithm1}, then for every $n,x$, 
\begin{equation}
\label{eq:the1}
        \pi^{k}_n(\cdot|x,{\mu_n^{k}})
        = \operatorname{softmax}\left(\frac{1}{\tau} \widetilde{Q}^k_n(x,{\mu_n^{k}},\cdot)\right).
\end{equation}
\end{theorem}

We emphasize that in Theorem~\ref{thm:equivalence}, the summation-form policy $\pi^{k-1}_n(\cdot|x,\mu)$ is used solely as an intermediate construct to derive the final, more concise policy representation $\pi^{k}_n(\cdot|x,\mu)$. The theorem establishes the equivalence between these two types of formulations. Hence, for every iteration $k$ in the algorithm, only the concise form given in equation~\eqref{eq:the1} is adopted as the policy representation.

\begin{proof}
The proof relies on the following two equalities:
$
     \operatorname{softmax}\big(\frac{1}{\tau}\sum_{i=0}^k Q^i\big)=\operatorname{argmax}_\pi\left\langle\pi, Q^k\right\rangle-\tau \mathrm{KL}\left(\pi \| \pi^{k-1}\right),
$
and:
$
\operatorname{softmax}\big(\frac{1}{\tau} \tilde{Q}^k\big) = \operatorname{argmax}_\pi\big\langle\pi, \tilde{Q}^k\big\rangle-\tau\langle\pi , \ln \pi\rangle $. We start by showing the two equalities hold, then prove the equivalence between the right-hand side of both equations. The left hand side of the first equality is the policy used in the original OMD-based ~\cite{perolat2021scaling} algorithm, while the second equality is the policy used in our algorithm. 

To prove the two equalities hold, we use the Lagrange multiplier method with the constraint that $\boldsymbol{1}^{\top} \pi=1$. For example, for the first equality, the problem is:

\begin{equation*}
\begin{gathered}
\underset{\pi}{\operatorname{argmax}} \, F(\pi)=\underset{\pi}{\operatorname{argmax}}\left\langle\pi, Q^k\right\rangle-\tau \mathrm{KL}\left(\pi \| \pi^{k-1}\right) \\
\text { s.t. } \quad \boldsymbol{1}^{\top} \pi=1,
\end{gathered}
\label{prob:optimization1}
\end{equation*}
where $\boldsymbol{1}$ denotes a vector full of ones, of dimension the number of actions. By solving this optimization with the Lagrange multiplier method, we can obtain the optimal policy as $\operatorname{softmax}\big(\frac{1}{\tau}\sum_{i=0}^k Q^i\big)$. Following the same way, we can prove the second equality.

After that, we now prove the equivalence between the two right-hand side. Let $\tilde{Q}^k=Q^k+\tau \ln \pi_{k-1}$.
Then 
\begin{equation*}
    \begin{aligned}
&\underset{\pi}{\operatorname{argmax}}\left\langle\pi \cdot Q^k\right\rangle-\tau \mathrm{KL}\left\langle\pi \| \pi^{k-1}\right\rangle 
\\
& =\underset{\pi}{\operatorname{argmax}}\left\langle\pi, \tilde{Q}^k-\tau \ln \pi_{k-1}\right\rangle-\tau \mathrm{KL}\left(\pi \| \pi^{k-1}\right) \\
& =\underset{\pi}{\operatorname{argmax}}\left\langle\pi, \tilde{Q}^k\right\rangle-\tau\langle\pi, \ln \pi\rangle
\end{aligned}
\end{equation*}

The detailed proof is shown in Appx. \ref{proof:softmax}.
\footnote{\tiny Long version with appendices available here: \url{https://drive.google.com/file/d/1nXKRwdVhSw-HogyzcnoGbz_9Pxw6wx3b/view?usp=sharing}}
\looseness=-1 

\end{proof}
Theorem \ref{thm:equivalence} suggests the update rule to be used. In the implementation, at iteration $k$ of our algorithm (see Algo.~\ref{algo: algorithm1}) we train a deep Q-network $\tilde{Q}^k_\theta$ with parameters $\theta$  which takes as inputs the time step, the agent's state, the mean-field state, and the agent's action. This DNN is trained to minimize the loss: 
$ \mathbb{E}\left|\tilde{Q}^k_\theta\left(\left(n, x_n,\mu_n\right), a_n\right) - T_n\right|^2,$ 
where the target $T_n$ is: 
\begin{equation}
\begin{split}
    T_n & =  \, r_n^k +  {\color{blue}L^{k}_n} + \\
    & \gamma \sum_{a_{n+1} \in \cA} \pi^{k}_{\theta^{\prime}}\left(a_{n+1} \mid s_{n+1}^k\right) \Big[
    \tilde{Q}^k_{\theta^{\prime}}\left(s_{n+1}^k, a_{n+1}\right) {\color{blue}-\, L^{k}_{n+1}}\Big]
\end{split}
\label{eq:target-Tn}
\end{equation}

with $r_n^i = r(x_n,a_n,\mu_n^k)$, $s_n^k = (n, x_{n}, \mu_{n}^k)$, and ${\color{blue}L^{k}_{n} = \tau \log (\pi^{k-1}_{\theta}\left(a_n \mid s_n^k \right))}$, which is the main difference with a classical DQN target.  
Here $\pi^{k}_{\theta^{\prime}} = \operatorname{softmax}(\frac{1}{\tau} \tilde{Q}^k_{\theta^{\prime}})$ where $\tilde{Q}^k_{\theta^{\prime}}$ is a target network with same architecture but parameters $\theta^{\prime}$, updated at a slower rate than $\tilde{Q}^k_{\theta}$. In the definition of $T_n$, the {\color{blue}terms in blue} involve $\pi^{k-1}_{\theta} = \operatorname{softmax}(\tfrac{1}{\tau} \tilde{Q}^{k-1}_{\theta})$, which performs implicit averaging .
Differing from the Q update function proposed in~\cite{lauriere2022scalable}, where the target policy is set as the policy learned from the previous iteration, our algorithm employs the target policy $\pi^{k}_{\theta^{\prime}}$ and target Q-function $\tilde{Q}^k_{\theta^{\prime}}$ being learned in the current iteration, namely $k$ instead of $k-1$. This modification helps stabilize the learning because if the target policy is fixed as a separate policy, the distribution of the policy under evaluation will differ from the one being learned. Consequently, this distribution shift would induce extra instability in the learning process.

\begin{algorithm} [htb]  
\caption{Master Deep Online Mirror Descent }\label{algo: algorithm1}
\SetAlgoLined
\SetKwData{Left}{left}\SetKwData{This}{this}\SetKwData{Up}{up}
\SetKwFunction{Union}{Union}\SetKwFunction{FindCompress}{FindCompress}
\SetKwInOut{Input}{input}\SetKwInOut{Output}{output}

\begin{flushleft}
  \textbf{Input:} Learning iteration ${K}$; training episodes ${N_{episodes}}$ per iteration; Replay buffer $\mathcal{M_{RL}}$; horizon $N_T$, number of agents ${N}$; Initial distribution $\mathcal{D}$; Munchausen parameter $\tau$, Initial Q-network parameter $\theta$; Initial target Q-network parameter $\theta^{\prime}$
\end{flushleft}

\begin{flushleft}
\textbf{Output:}  {Policy $\boldsymbol{\pi}$}
\end{flushleft}

\begin{flushleft}
Set initial target network parameter $\theta^{\prime}$ = $\theta$ 

Set initial 
$\boldsymbol{\pi}^0(a \mid(n, x,\mu))=\operatorname{softmax}\left(\frac{1}{\tau} {Q}_{\theta}((n, x,\mu), \cdot)\right)$
\end{flushleft}
\begin{flushleft}
\For{iteration $k=1,2,\dots,K$} 
{
1. Update mean-field sequence:

\For{ distribution ${\bsmu^k}$ in $(\bsmu^{k,\mu_0})_{\mu_0 \in \mathcal{D}}$}
{
Update mean-field sequence $\bsmu^{k}$ with $\boldsymbol{\pi}^{k-1}$ sampled by agents ${N}$ 
}
2. Reset the replay buffer $\mathcal{M_{RL}}$ 

3. Value function update:

\For{episode $t=1,2,\dots,N_{episodes}$}
 {
 \For{  ${\bsmu^k}$ in $(\bsmu^{k,\mu_0})_{\mu_0 \in \mathcal{D}}$}
 {
     \For{time step $n=1,2,\dots,N_T$}
        {
            Sample action $a_{n}$ from $\epsilon$-greedy policy based on $\tilde{Q}_{\theta}$ 
            
            Execute action $a_{n}$ and get the transition: 
            $\left\{\left(\left(n, x_{n}, \mu_{n}^k\right), a_n, r_n,\left(n+1, x_n^{\prime}, \mu_{n+1}^k\right)\right)\right\}$
            
            Store transition in replay buffer $\mathcal{M_{RL}}$ 
    
            Periodically update $\theta$ with one step gradient step using a minibatch $N_B$ from $\mathcal{M_{RL}}$: 
            $\theta \mapsto \frac{1}{N_B} \sum_{i=1}^{N_B}\left|\tilde{Q}_\theta\left(\left(n_i, x_{n_i},\mu_{n_i}^k\right), a_{n_i}\right)-T_{n_i}\right|^2$
            
            where $T$ is defined in \eqref{eq:target-Tn} 
            
            Periodically update target network parameter $\theta^{\prime} = \theta$
        }
 }
 
 }
    
4. Policy update: 
$\boldsymbol{\pi}^k(\cdot \mid(n, x,\mu))=\operatorname{softmax}\left(\frac{1}{\tau} \tilde{Q}_{\theta}((n, x,\mu), \cdot)\right)$
} 
Return $\boldsymbol{\pi}^K$
\end{flushleft}
\end{algorithm}

\paragraph{Inner loop replay buffer. }
To effectively learn a master policy capable of handling any initial distribution, this paper leverages the concept of replay buffer \cite{mnih2015human, schaul2015prioritized} and uses it to mix knowledge from different initial distributions. In the MFG setting considered here, the states incorporate both the distribution and timestep. Thus, using a replay buffer aligns well with stationary sampling requirements. However, maintaining the buffer for the entire history imposes substantial computational overhead and slows learning. Additionally, if the DNN is trained on multiple initial distributions, separate evolutionary processes must be inputted. This approach risks catastrophic forgetting \cite{kirkpatrick2017overcoming}, where learning from earlier processes may be lost if the gap between them is too wide.
Our algorithm addresses these challenges by utilizing the implicit summation of historical Q-values, as formalized in Thm~\ref{thm:equivalence}. This approach eliminates the need to sample from earlier iterations during subsequent training. As a result, we limit the replay buffer size and reset it at the start of each iteration $k$. This efficient buffer management strategy, detailed in Step 2 of Algo.~\ref{algo: algorithm1}, is supported by empirical results (see Fig.~\ref{fig:buffer_replay}).

\section{MFG with Common Noise}   
\label{sec:common-noise}
Our approach to learn master policies can handle common noise. In the context of MFGs, \defi{common noise} (also called aggregate shock) refers to randomness that affects all players \cite{carmona2016mean}. We denote it by \(\Xi_N = \{\xi_n\}_{0 \leq n \leq N} = \Xi_{N-1} \cdot \xi_{N}\), and it influences transitions and rewards. Consequently, population distributions and policies are conditioned on this noise, written as \(\pi_n(a \mid x, \mu_n, \Xi_n)\) and \(\mu_n(a \mid x, \Xi_n)\), respectively. The Q-function of a representative player is defined as:
\[
\begin{cases}
     Q_{N_T}(x,\mu,a,\Xi_{N_T})= r_{N_T}(x,\mu,a,\xi_{N_T}),
    \\
    Q_n(x, \mu, a,\Xi_{n}) = r_n(x, \mu, a, \xi_n) +  \\
    \quad \quad \quad 
    \mathbb{E}_{x',a',\xi_{n+1}}\left[Q_{n+1}(x', \mu', a',\Xi_{n+1}) \right],  n \in [N_T-1]
\end{cases} 
\]
\text{where } $x' \sim p_n(x' \mid x, \mu, a, \xi_n),$ $\mu' = \sum_{x,a} \mu(x)\pi_n(a|x,\mu,\Xi_{n}) p_n\left(x' \mid x, a, \mu, \xi_n\right)$ and $a' \sim \pi_{n+1}(\cdot \mid x',\mu ', \Xi_{n+1})$.

Exploitability is defined as in \eqref{eq:exploitabiliy}, with policies and distributions conditioned on the common noise. Following \cite{perolat2021scaling}, the proof of convergence relies on constructing a similarity function that measures the proximity between current policies and the Nash equilibrium. The gradient of this function has two components: the first relates to exploitability, and the second corresponds to the monotonicity condition, namely,
\[
    \textstyle
    \sum_{x \in \mathcal{X}}\left(\mu(x|\xi)-\mu^{\prime}(x|\xi)\right)\left(\bar{r}(x, \mu,\xi)-\bar{r}\left(x, \mu^{\prime},\xi\right)\right) \leq 0,
\] 
where $\bar{r}$ is the reward of interaction with the population. Under this condition, the gradient of the similarity function will be non-positive until it reaches the Nash equilibrium.

Our master policies, trained with DNNs, naturally extend to incorporate common noise. While Q values, policies, and distributions depend on the noise, distinguishing continuous changes in practice is challenging. Thus, the policy and Q-network input include the entire historical sequence $\Xi_n$ up to the current step, rather than just the current state $\xi_n$.

\section{Experimental results}
\label{sec:expe}
\subsection{Experimental setup}
 \textbf{Environments.} 
We consider seven examples in three environments that are canonical benchmarks for MFG domains. In MFGs, solving games requires identifying an equilibrium, which is more complex than reward maximization. Additionally, the mean-field evolution depends on the initial distribution. Each experiment explores two scenarios. The first, referred to as {\bf fixed $\mu_0$}, follows the common practice of starting the population from a fixed initial distribution. The second, {\bf multiple $\mu_0$}, tests the master policy's effectiveness by using various initial distributions during training. Detailed distributions are provided in the Appx. Unlike training separate networks for different Nash equilibria, our master policy uses a single network to learn equilibrium policies for all initial distributions. Population-independent policies typically underperform in this scenario, except when equilibrium policies remain unchanged across initial distributions, indicating no interactions. To illustrate the influence of common noise, we include the 1D Beach Bar and Linear Quadratic (LQ) examples.

\paragraph{Algorithms.} We compare our algorithm to four baselines, including several SOTA DRL algorithms for MFGs. In figures and tables, {\bf vanilla FP (V-FP)} represents an adaptation of the tabular FP from \cite{perrin2020fictitious} to DNNs. V-FP uses classic fictitious play (Algo.~\ref{algorithm2}) to iteratively learn Nash equilibrium, assuming agents always start from a fixed distribution. {\bf Master FP (M-FP)}, from \cite{Perrin2022General}, handles any initial distribution via FP. {\bf Vanilla OMD1 (V-OMD1)}, based on the Munchausen trick, is introduced in \cite{lauriere2022scalable}. {\bf Vanilla OMD2 (V-OMD2)} is our algorithm without the mean-field state as input, while the full version is called {\bf Master OMD (M-OMD)}. Both M-FP and M-OMD learn population-dependent policies, while V-FP, V-OMD1, and V-OMD2 do not. V-OMD2 serves as an ablation study of M-OMD, excluding distribution dependence to test its impact.

\paragraph{Implementation. } FP-type algorithms use the DQN algorithm to learn the best response to the current mean-field sequence iteratively. In the model-free setting, transition probabilities and reward functions are unknown during training and execution. Our Q-network follows the DQN architecture \cite{mnih2015human, ota2021training}. Similarly, OMD-type algorithms use a comparable Q-network for policy evaluation. Distributions are represented as histograms, converted into one-dimensional vectors (or concatenated if needed) before passing to the network. This approach works well for the studied examples. For 2D cases, ConvNets could improve performance \cite{Perrin2022General}, though they were unnecessary in our experiments. Importantly, our method learns \emph{non-stationary} policies, which are critical for finite-horizon problems. Unlike infinite-horizon settings \cite{Perrin2022General}, timesteps are essential here and are incorporated into the agent's policy using one-hot encoding. Here, we must explain the reason for using one-hot encoding: The reason for feeding the network with one-hot encoded timesteps instead of embedding real-valued timesteps is that our system operates in discrete time. The numerical values between two successive timesteps, such as 
$n$ and $n-1$, have no meaningful interpretation. If real-valued time were used as input, the network might mistakenly assume that there are meaningful values between adjacent timesteps, leading to ineffective learning.

For common noise, we pre-generate the sequence, progressively reveal it during training, and pad zeros for unobserved timesteps. This keeps the input length constant while using only available noise data. To visualize performance, we implemented model-based methods and fine-tuned hyperparameters for each algorithm, using the best parameters found.

The GPU used is NVIDIA TITAN RTX (24gb), the CPU is 2x 16-core Intel Xeon Gold (64 virtual cores). The exploitability curves are averaged over 5 realizations of the algorithm, and whenever relevant, we show the standard deviation (std dev) with a shaded area. The details about training and hyperparameters are listed in the Appx..

\subsection{Env1: Exploration}
\label{sec:expe:exploration}
Exploration is a classic problem in MFG \cite{geist2021concave}, where a large group of agents tries to avoid crowded areas and hence uniformly distribute into empty areas in a decentralized way. Here we introduce two variants, with different geometries of domain: exploration in one room, and four connected rooms, which is more challenging.

\begin{figure}[htb]
\centering
\subfloat[Evolution process]{
\begin{minipage}[t]{0.5\linewidth}
\centering
\includegraphics[width=1.5in]{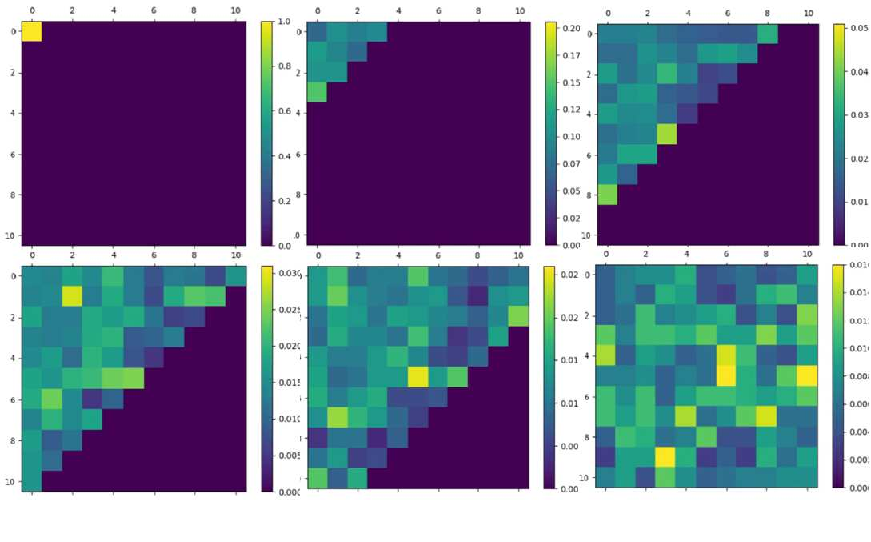}
\end{minipage}%
}%
\subfloat[Exploitability (fixed $\mu_0$)]{
\begin{minipage}[t]{0.5\linewidth}
\centering
\includegraphics[width=1.5in]{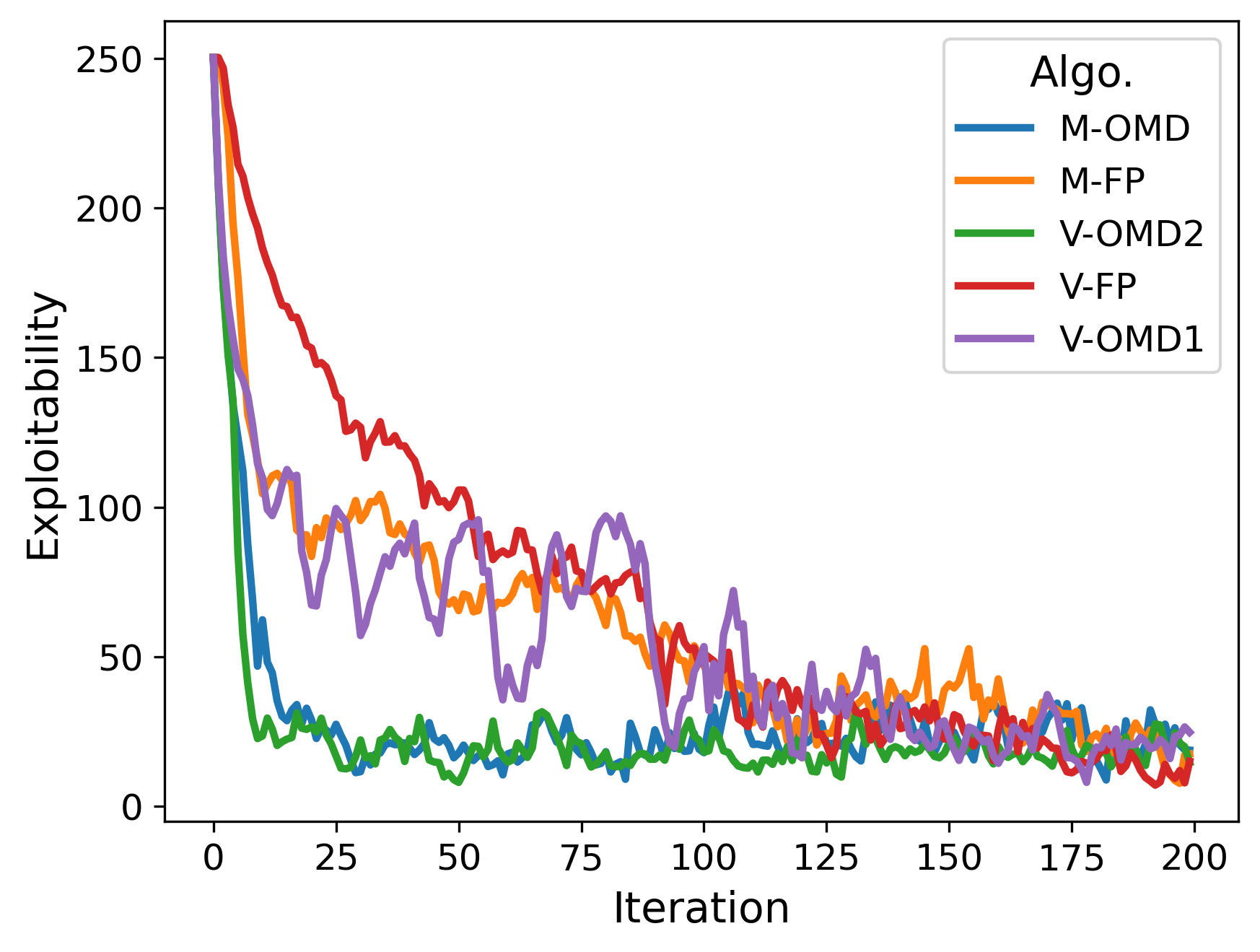}
\end{minipage}%
}%
\newline
\subfloat[Exploitability (multiple $\mu_0$)]{
\begin{minipage}[t]{0.5\linewidth}
\centering
\includegraphics[width=1.5in]{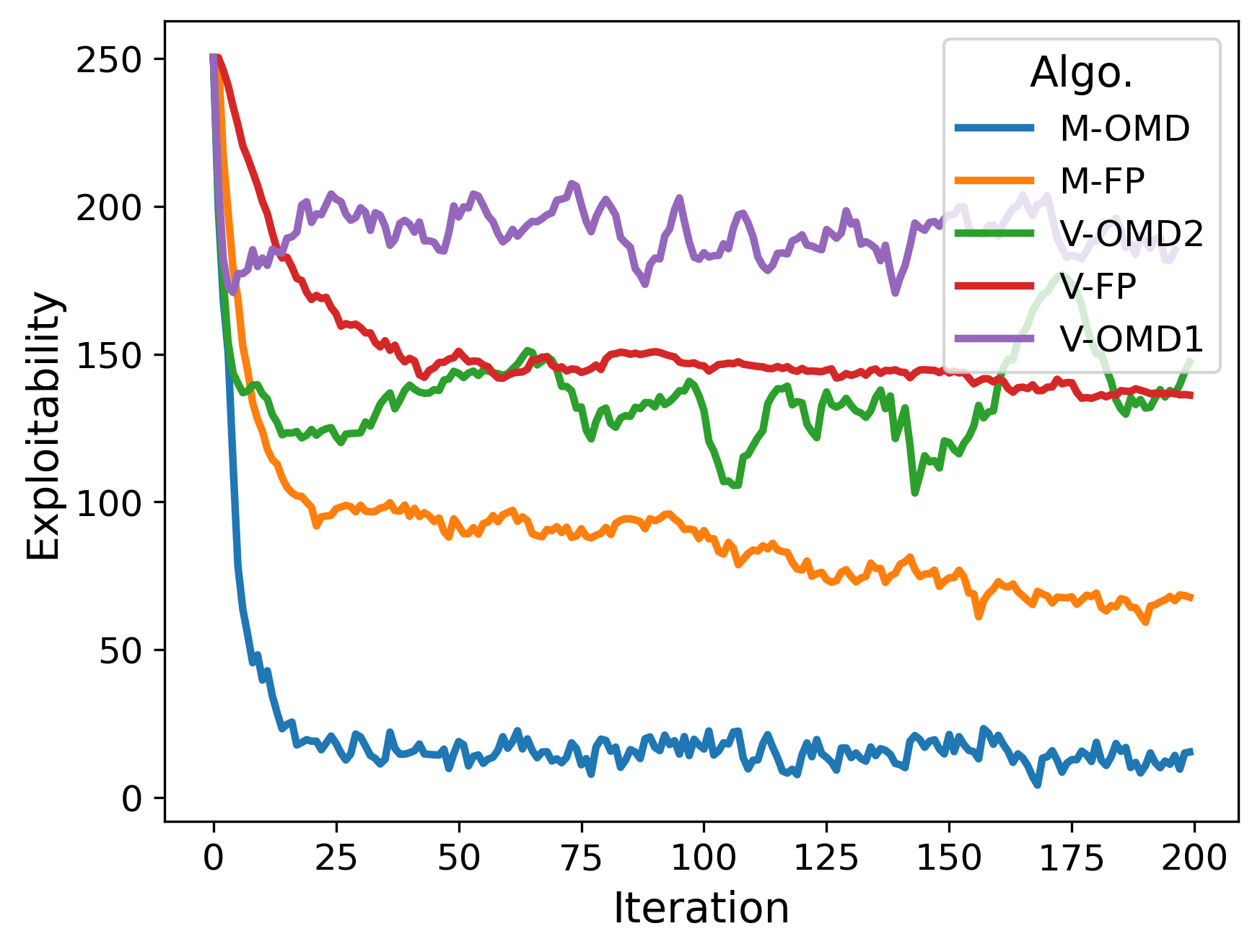}
\end{minipage}%
}%
\subfloat[Exploitability (multiple $\mu_0$)]{
\begin{minipage}[t]{0.5\linewidth}
\centering
\includegraphics[width=1.5in]{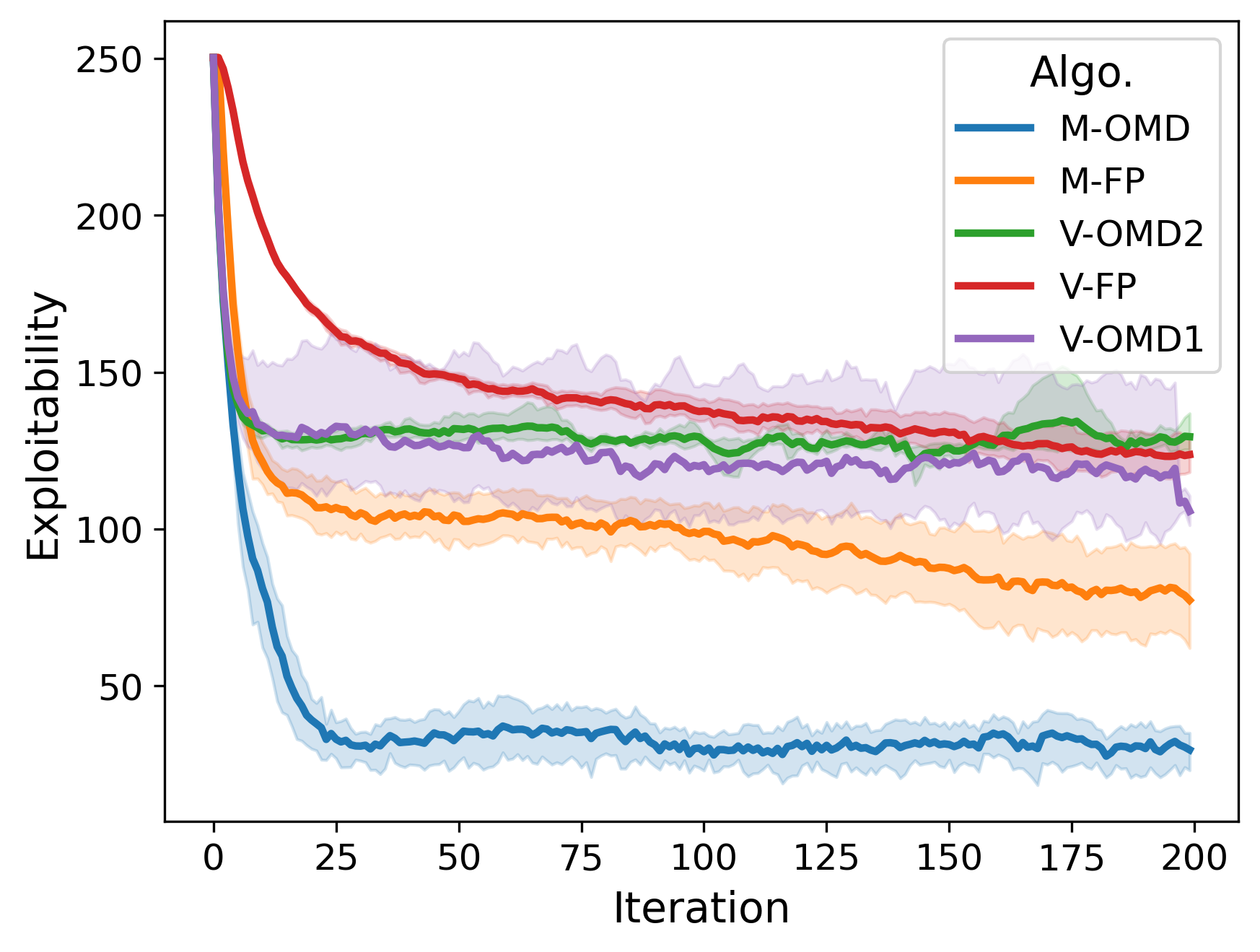}
\end{minipage}
}%
\centering
\caption{Example 2 in Env1: Exploration in one room. (a): density evolution using the policy learnt by M-OMD, starting from the $\mu_0$ used for (b). (b): exploitability vs training iteration for a single $\mu_0$. (c): average exploitability when training over 5 different $\mu_0$ (single run of each algo.). (d): averaged curve over 5 runs and std dev.}
\label{uniform_fig}
\end{figure}

\paragraph{Example 1: Exploration in One Room}
We consider a 2D  grid world of dimension $11 \times 11$. The action set is $\mathcal{A}$=\{up, down, left, right, stay\}. The dynamics are:
$    x_{n+1}=x_n+a_n+\epsilon_n,$ 
where $\epsilon_n$ is an environment noise that perturbs each agent's movement (no perturbation w.p. $0.9$, and one of the four directions w.p. $0.025$ for each direction). 
The reward function will discourage agents from being in a crowded location:
 $   
 r(x, a, \mu)=-\log (\mu(x))-\frac{1}{|X|}|a|.
 $ The result is shown in Fig. \ref{uniform_fig}.

\paragraph{Example 2: Exploration in four connected rooms. }
This environment consists of four connected rooms where agents cannot move through obstacles. The goal is to explore every grid point within the map, which has dimensions $11\times11$. The action set is $\mathcal{A}$=\{up, down, left, right, stay\}. The dynamics are: $    x_{n+1}=x_n+a_n+\epsilon_n,$, where $\epsilon_n$ is an environment noise that perturbs each agent's movement (no perturbation w.p. $0.9$, and one of the four directions w.p. $0.025$ for each direction). 
The reward function will discourage agents from being in a crowded location:
 $   
 r(x, a, \mu)=-\log (\mu(x))-\frac{1}{|X|}|a|.
 $

\begin{figure}[htb]
\centering
\subfloat[Evolution process]{
\begin{minipage}[t]{0.5\linewidth}
\centering
\includegraphics[width=1.5in]{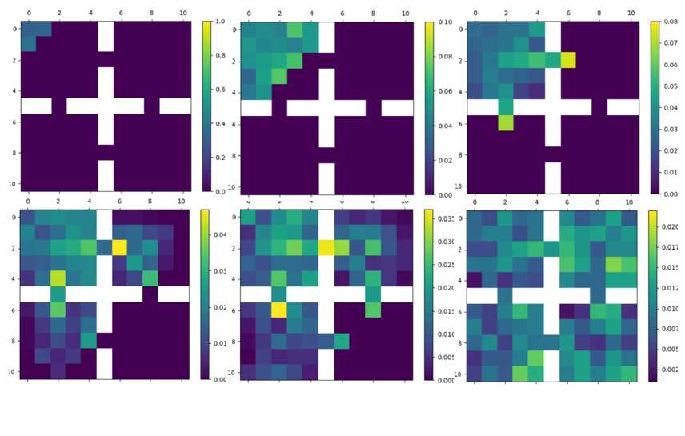}
\end{minipage}%
}%
\subfloat[Exploitability (fixed $\mu_0$)]{
\begin{minipage}[t]{0.5\linewidth}
\centering
\includegraphics[width=1.5in]{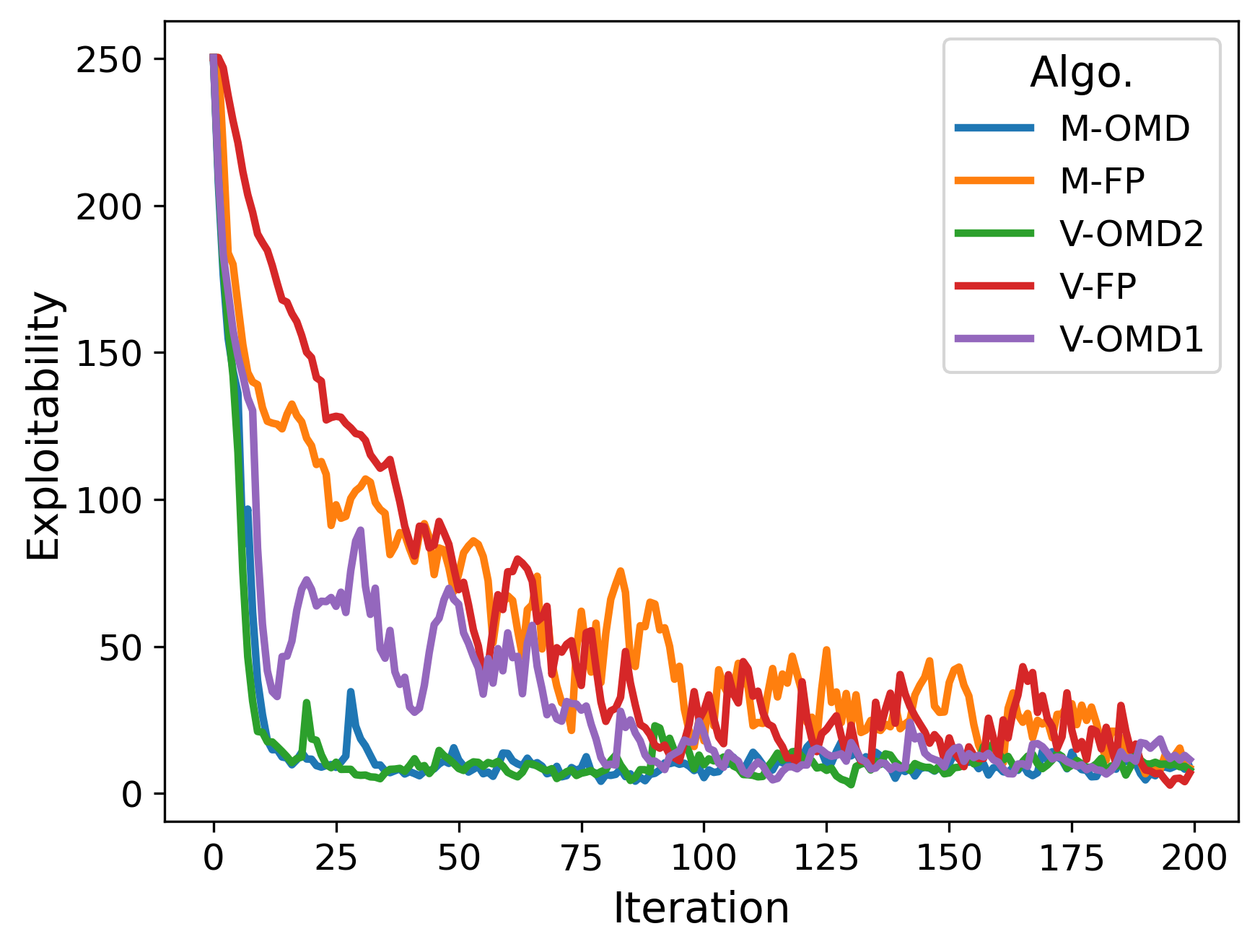}
\end{minipage}%
}%
\newline
\subfloat[Exploitability (multiple $\mu_0$)]{
\begin{minipage}[t]{0.5\linewidth}
\centering
\includegraphics[width=1.5in]{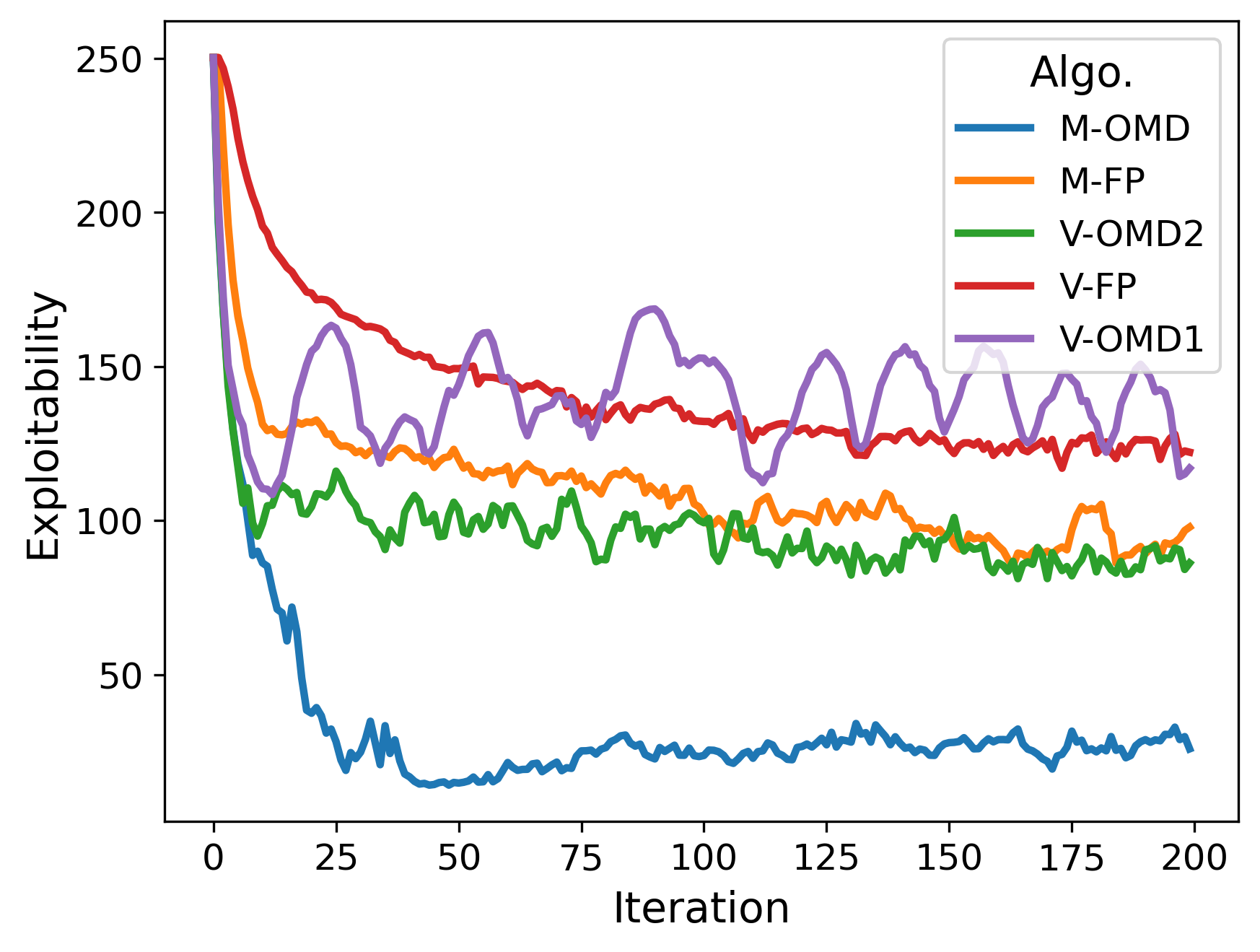}
\end{minipage}%
}%
\subfloat[Exploitability (multiple $\mu_0$)]{
\begin{minipage}[t]{0.5\linewidth}
\centering
\includegraphics[width=1.5in]{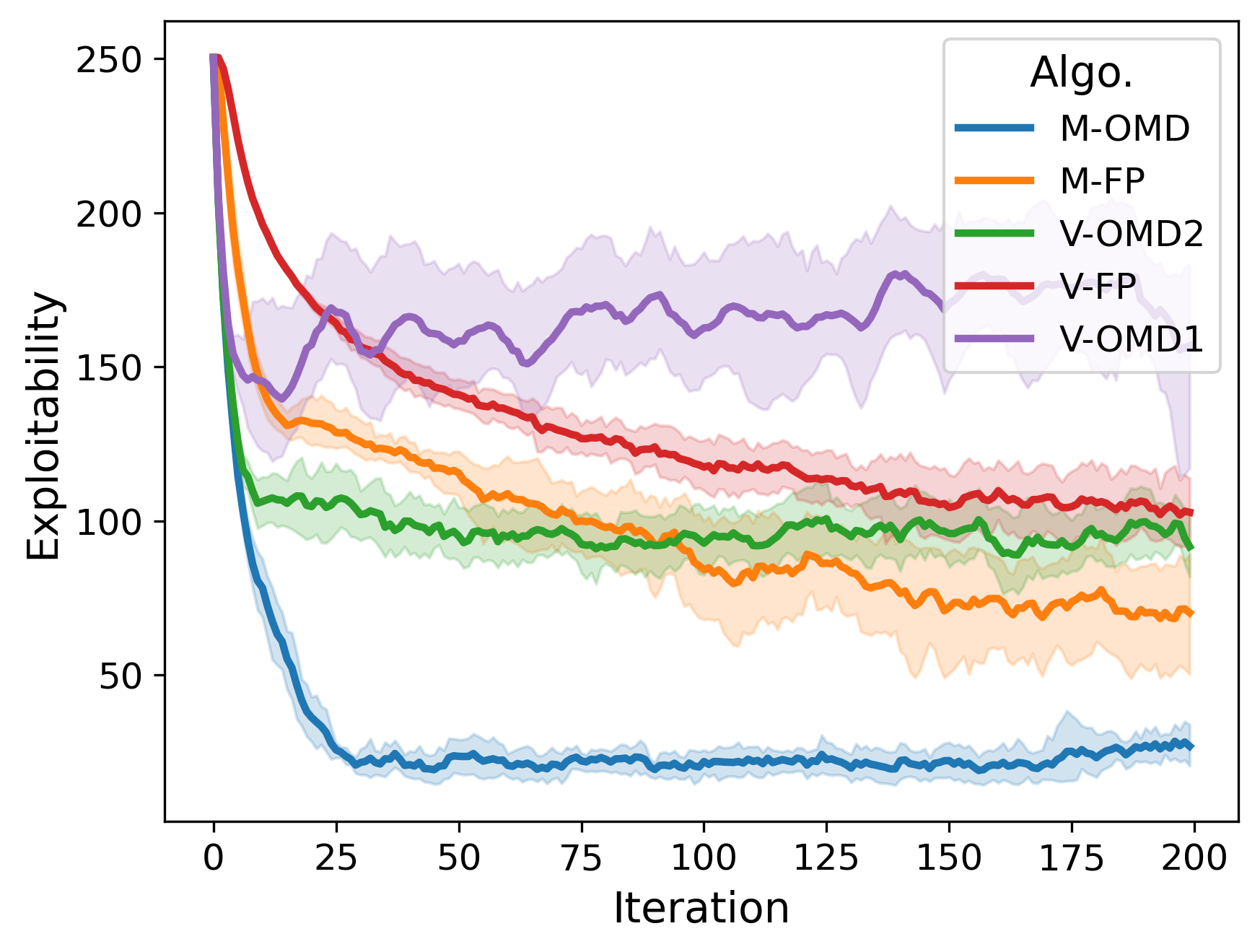}
\end{minipage}
}%
\centering
\caption{Example 1 in Env1: Exploration in four connected rooms. 
(a): density evolution using the policy learned by M-OMD, starting from the $\mu_0$ used for (b). (b): exploitability vs training iteration for a single $\mu_0$. (c): average exploitability when training over 5 different $\mu_0$ (single run of each algo.). (d): average over 5 runs \& std dev.
}
\label{exploration_fig}     
\end{figure}

Fig.~\ref{exploration_fig} shows the results (see Fig.~\ref{uniform_fig} for the one-room geometry): heat-maps representing the evolution of the distribution (at several time steps) when using the master policy learned by our algorithm; evolution exploitability when using a fixed $\mu_0$ or multiple $\mu_0$ for a single run of the algorithm; and finally the results averaged over 5 runs. On both examples, our proposed algorithm (M-OMD) converges faster than all the 4 baselines. With fixed $\mu_0$, all methods perform well, but with multiple initial distributions, it appears clearly that F-FP, V-OMD1 and V-OMD2 fail to converge, due to the fact that vanilla policies lack awareness of the population so agents cannot adjust their behavior suitably when the initial distribution varies. In other words, \emph{population-independent policies cannot be Nash equilibria when testing on new initial distributions}, hence their exploitability is non-zero. 

\subsection{Env2: Beach bar}
The Beach bar environment, introduced in~\cite{perrin2020fictitious} represents agents moving on a beach towards a bar. The goal for each agent is to avoid the crowd but get close to the bar. The dynamics are the same as in the exploration examples. Here we consider that the bar is located at the center of the beach, and it is not possible to go beyond the domain. The beach bar without common noise is shown in Fig. \ref{fig:beach_bar} in a $11\times11$ (2D) map.

\begin{figure}[htb]
\centering
\subfloat[Model-based Evolution]{
\begin{minipage}[t]{0.5\linewidth}
\centering
\includegraphics[width=1.5in]{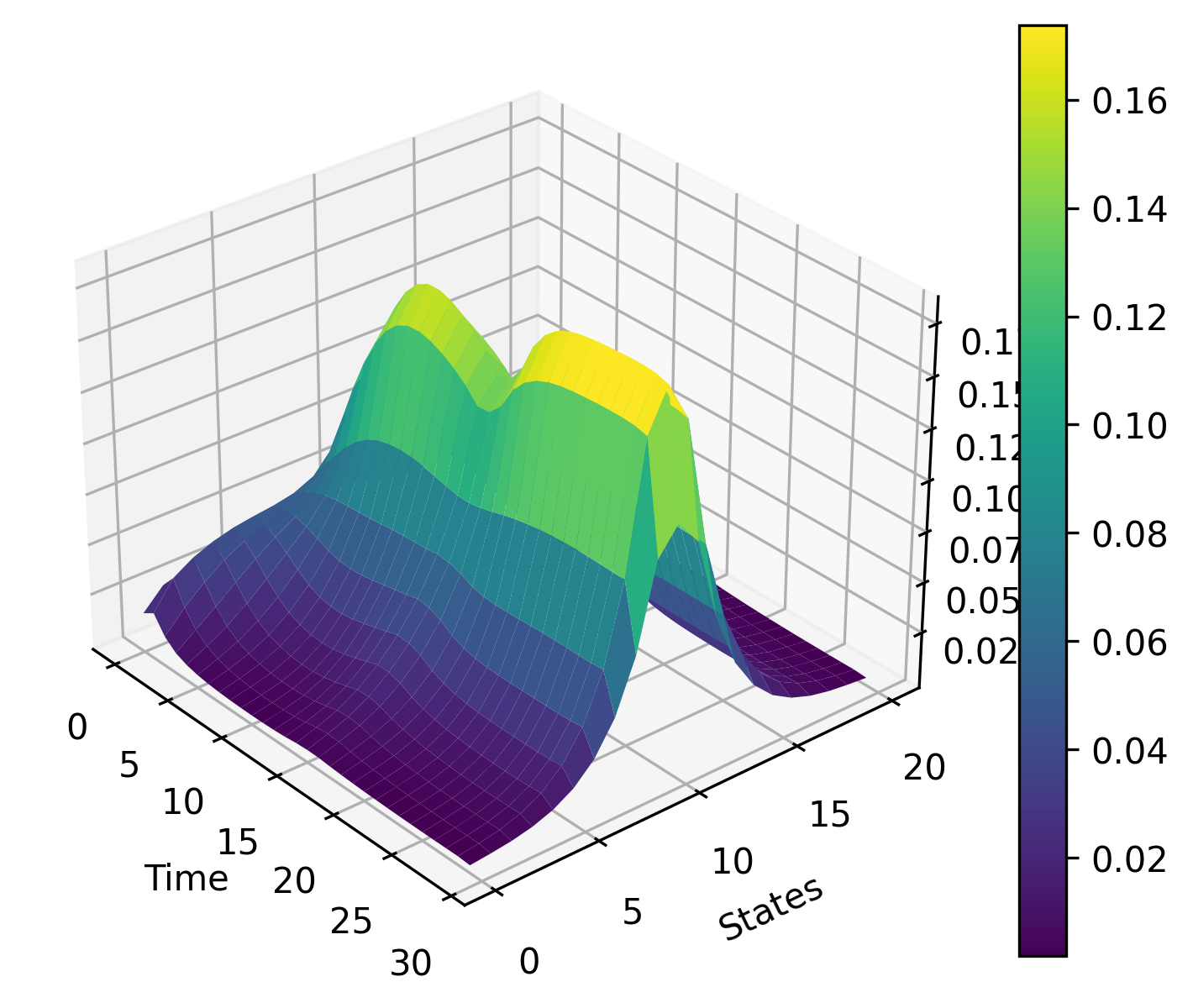}
\end{minipage}
}%
\subfloat[Master-FP Evolution]{
\begin{minipage}[t]{0.5\linewidth}
\centering
\includegraphics[width=1.5in]{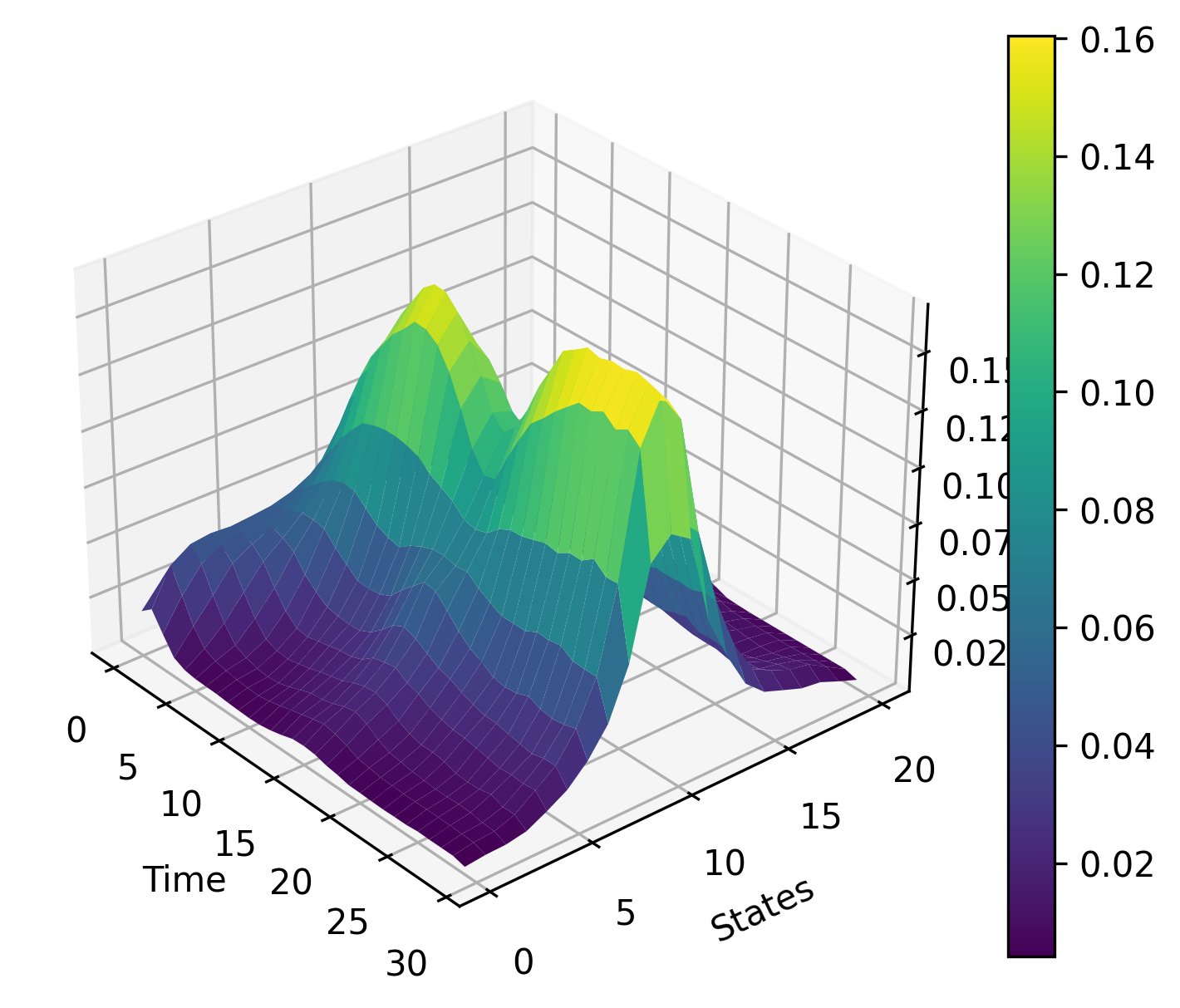}
\end{minipage}
}\
\subfloat[Master-OMD Evolution]{
\begin{minipage}[t]{0.5\linewidth}
\centering
\includegraphics[width=1.5in]{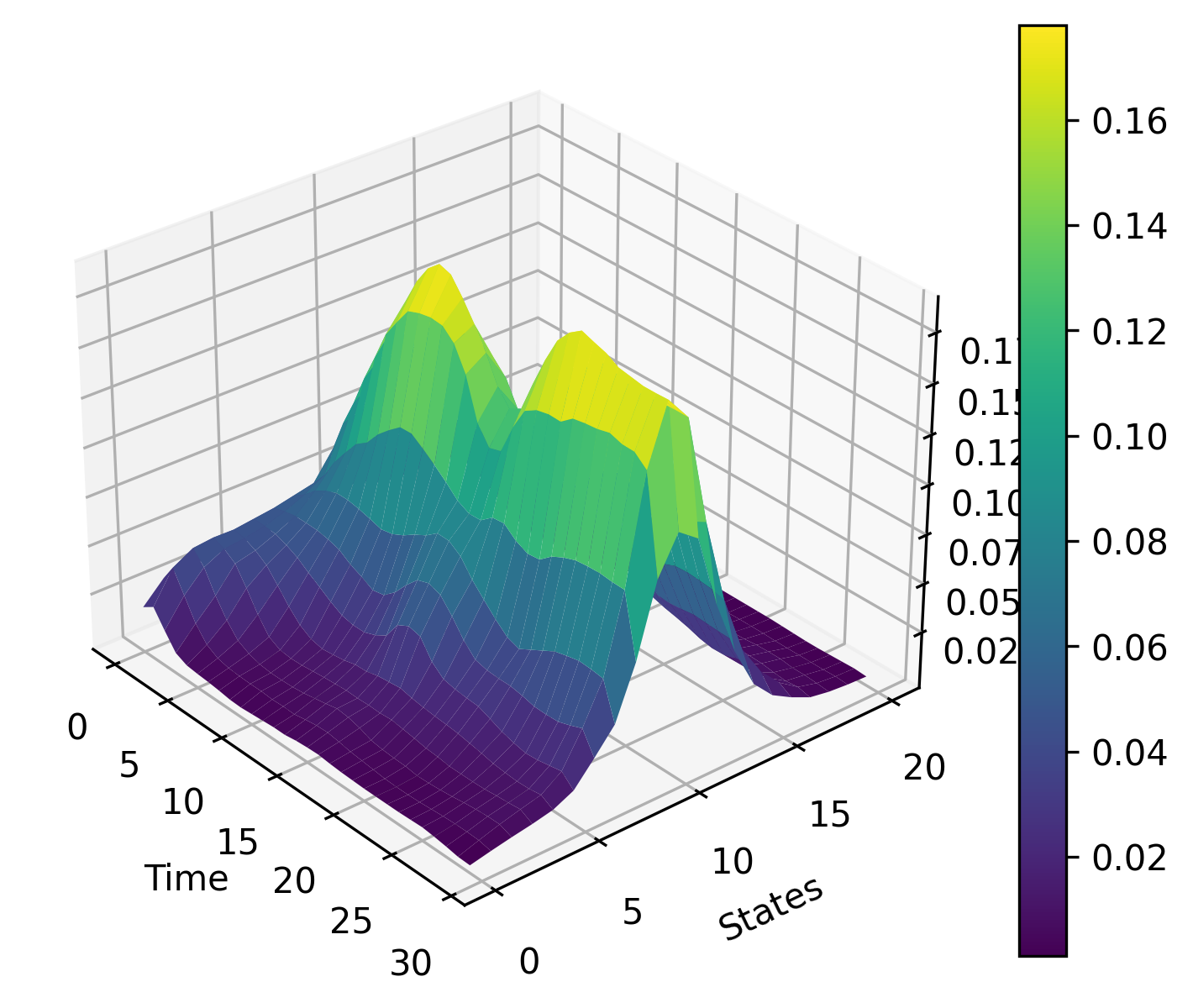}
\end{minipage}
}%
\subfloat[Exploitability]{
\begin{minipage}[t]{0.5\linewidth}
\centering
\includegraphics[width=1.5in]{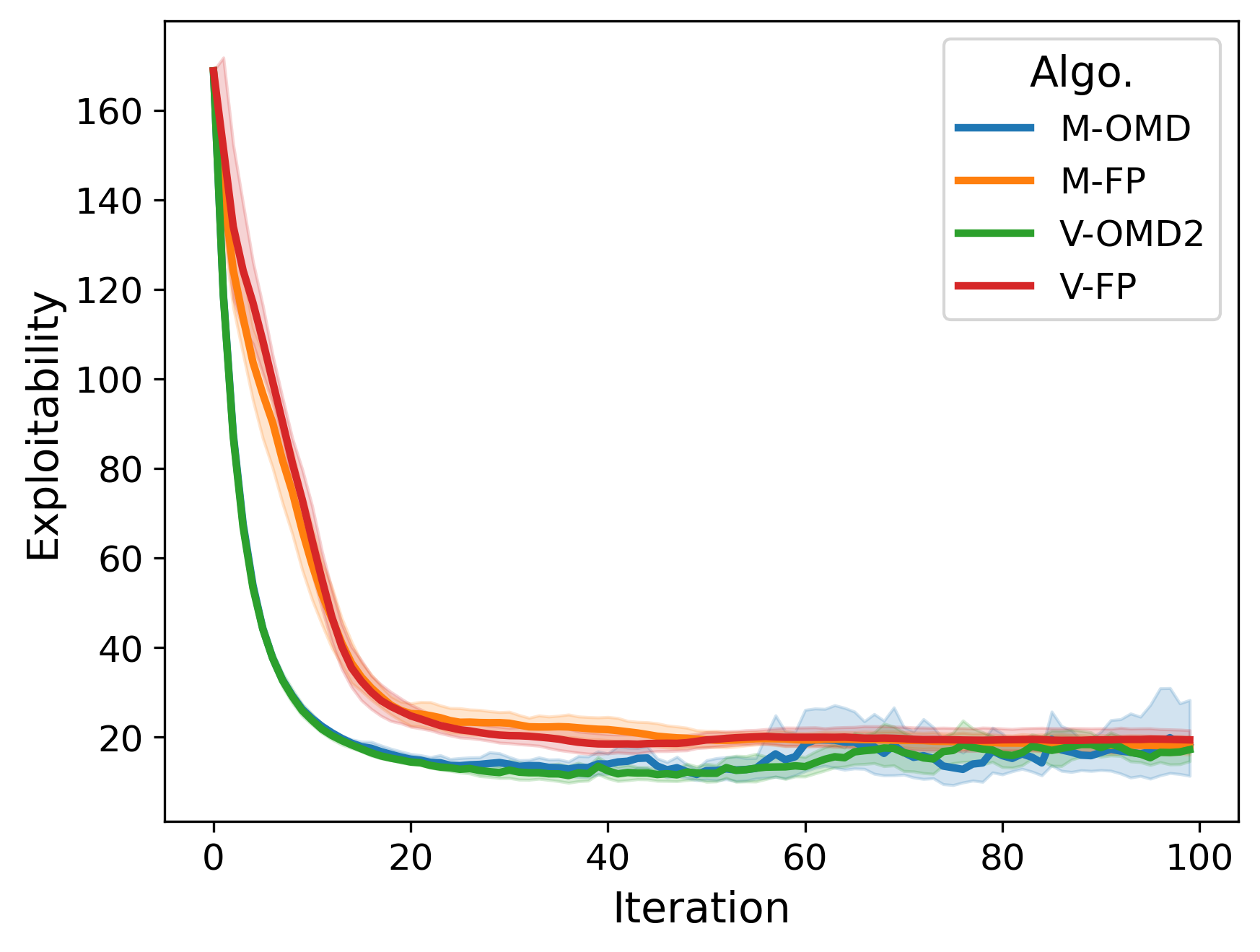}
\end{minipage}
}\

\caption{Example 1 in Env2: Beach bar with the ``Closure'' Common Noise. The population has no prior information about the status of the bar. (a) shows the model-based solution, (b) and (c) present the master FP and master OMD solutions, respectively. (d) shows the exploitability of multiple seeds.}
\label{fig:beach_bar_cn1}
\end{figure}

\paragraph{Example 1: Beach bar with Common Noise }
The reward function is:
$
r\left(x, a, \mu\right)={d_{bar}}\left(x\right)-\frac{\left|a\right|}{|\mathcal{X}|}- C \log \left(\mu\left(x\right)\right), 
$
where $d_{bar}$ represents the distance to the bar, the second term penalizes movement to encourage minimal action, and the third term discourages agents from occupying crowded regions. The parameter $C=1$  when the bar is open and $C=0$  when closed. The domain $\mathcal{X}$ is 1D with $11$ (1D) states. The common noise follows \cite{perrin2020fictitious} and acts as a random switch determining whether the bar is open or closed. The bar's status changes randomly midway through the game. Figure~\ref{fig:beach_bar_cn1} visualizes the dynamic evolution of the master FP and master OMD policies compared with the model-based solution under this "closure" noise scenario.

\subsection{Env3: Linear-Quadratic with Common Noise}
\label{sec:Env3-Ex1-CN}
The linear-quadratic (LQ) model is a classical setting studied e.g. in~\cite{carmona2013control,bensoussan2016linear}. A discretized version was introduced in~\cite{perrin2020fictitious}. We present results with common noise here; results without common noise are in Appx.~\ref{app:lq}.

\paragraph{Example 1: LQ with Common Noise }
The LQ with common noise is a 1D model, in which the dynamics are: $x_{n+1} = x_n + a_n \Delta_n + \sigma \left( \rho \xi_n + \sqrt{1 - \rho^2} \epsilon_n \right) \sqrt{\Delta_n}$, corresponding to moving by a number of states (left or right). The state space is $\mathcal{X}=\{-L, \ldots, L\}$, of dimension $|\mathcal{X}|$ =$2L-1$. To add stochasticity into this model, $\epsilon_n$ is an additional noise will perturb the action choice with $\epsilon_n \sim \mathcal{N}(0,1)$, but was discretized over $\{-3 \sigma, \ldots, 3 \sigma\}$. The reward function is:
\[
    r_n\left(x,a,\mu\right)=\big[-\tfrac{1}{2}\left|a\right|^2+q a\left(m-x\right)-\tfrac{\kappa}{2}\left(m-x\right)^2\big] \Delta_n,
\]
where $m=\sum_{x \in \mathcal{X}} x \mu(x)$ represents the population mean, encouraging agents to align with the population's center while maintaining dynamic movement. The terminal reward is $r_{N_T}\left(x, a, \mu\right)=-\frac{c_{\text {term }}}{2}\left(m-x\right)^2$. Here we used $\sigma=1$, $N_T=30$, $\Delta_n=1$, $q=0.01$, $\kappa=0.5$, $K=1$ $M=3$, $L=50$, $c_{term}=1$, \(\xi_n\) has two instances as:
\[
\begin{aligned}
\xi_n^1 &=
\begin{cases} 
-10 & n \leq 8 \\ 
0 & 8 < n \leq 20 \\ 
10 & n > 20 
\end{cases}
\quad
\xi_n^2 &=
\begin{cases} 
10 &  n \leq 8 \\ 
0 &  8 < n \leq 20 \\ 
-10 &  n > 20 
\end{cases}
\end{aligned}
\]

\begin{figure}[htb]
\centering
\subfloat[Model-based Evolution]{
\begin{minipage}[t]{0.45\linewidth}
\centering
\includegraphics[width=1.5in]{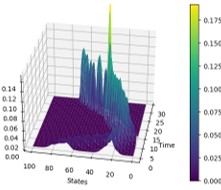}
\end{minipage}
}%
\subfloat[Master-FP Evolution]{
\begin{minipage}[t]{0.45\linewidth}
\centering
\includegraphics[width=1.5in]{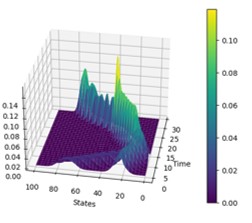}
\end{minipage}
}\
\subfloat[Master-OMD Evolution]{
\begin{minipage}[t]{0.45\linewidth}
\centering
\includegraphics[width=1.5in]{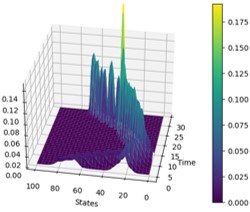}
\end{minipage}
}%
\subfloat[Exploitability]{
\begin{minipage}[t]{0.45\linewidth}
\centering
\includegraphics[width=1.5in]{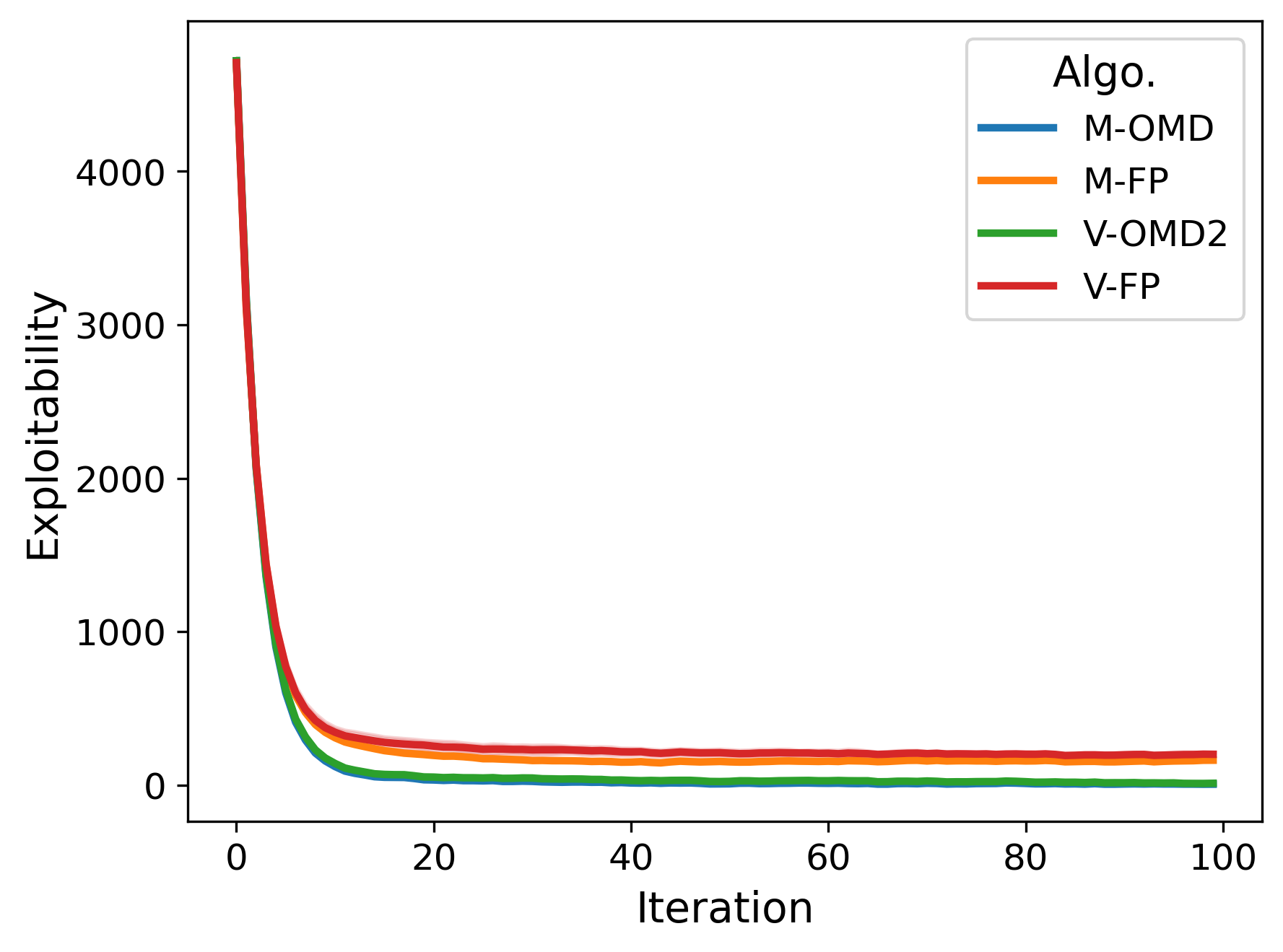}
\end{minipage}
}
\caption{Example 1 in Env3: Linear Quadratic with the $\xi_n^1$ type noise. The population only perceives the wave (i.e. disturbance) up to the present moment. (a) shows the model-based solution, (b) and (c) present the master FP and master OMD solutions, respectively. (d) shows the exploitability of multiple seeds.}

\label{fig:lq_cn1}
\end{figure}


This common noise creates bell-shaped disturbances in the population, mimicking effects like water flow on fish schools. Figures \ref{fig:lq_cn1} and \ref{fig:lq} illustrate that common noise significantly influences population trajectories. While agents converge toward the population center, the overall trajectory aligns with the noise direction. Theoretically, agents adapt to disturbances to maximize rewards, avoiding unnecessary energy expenditure to counteract the noise. This alignment between theory and experiment highlights the robustness of the approach.

\section{Discussion}
\label{sec:discu}

{\bf Numerical results. } We provide heuristic comparisons between the studied algorithms. Compared with V-OMD1 \cite{lauriere2022scalable}, V-OMD2 (ours) differs in three key aspects. First, it avoids convergence issues by not reusing the old policy as both behavior and target policy. Second, it employs a stabilized clip threshold $10^{-6}$ to avoid singularities, instead of another individual hyperparameter. Third, it incorporates an inner-loop replay buffer to mitigate catastrophic forgetting. Compared to M-FP \cite{Perrin2022General}, our algorithm achieves faster convergence, likely due to its constant update rates and implicit regularization via the Munchausen trick, avoiding FP-based issues (decaying rates and averaging over past iterations). Additionally, M-OMD demonstrates greater memory efficiency, as shown in Fig.~\ref{memory_cost} in Appx.
Furthermore, M-OMD is very flexible and successfully handles more complex problems, such as the exploration problem with \emph{ad-hoc teaming} (see Appx.~\ref{sec:adhoc-teaming}). This variant involves randomly introducing new agents mid-game, simulating real-life scenarios. Finally, Table~\ref{table:exploitability_testing} highlights that M-OMD consistently outperforms state-of-the-art baselines, including M-FP, except in the LQ case, where equilibrium policies appear nearly population-independent under the studied conditions.

\begin{table}[htb]
\begin{center}
\begin{scriptsize} 
\begin{sc}
\setlength\tabcolsep{5pt} 
\begin{tabular}{lccccc}
\toprule
Env. & V-FP & M-FP & V-OMD1 & V-OMD2 & \textbf{M-OMD}\\
\midrule
Exploration-1 & 135.32 & 84.76 & 175.91 & 163.42 & \textbf{78.2} \\
Exploration-2 & 146.45 & 72.66 & 166.84 & 159.67 & \textbf{60.0} \\
Beach Bar & 83.24 & 40.12 & 80.81 & 61.71 & \textbf{22.05} \\
LQ & 0 & 22.78 & 0 & 0 & \textbf{8.20} \\
\bottomrule
\end{tabular}
\end{sc}
\end{scriptsize}
\end{center}
\caption{Exploitability in testing set (200 training iterations)}
\label{table:exploitability_testing}
\end{table}

\noindent
{\bf Related works. } To the best of our knowledge, M-OMD is the first method to handle common noise and unknown initial distribution in finite-horizon MFGs. Most works on RL for MFGs focus on population-independent policies, see e.g.~\cite{guo2019learning,cui2021approximately}. \cite{lauriere2022scalable} combines FP and OMD with DRL but still for population-independent policies. \cite{Perrin2022General} learns master policies but relies on FP, while we rely on OMD, which is faster as shown in our experiments.

\section{Conclusion}
\label{sec:conclusion}
This paper presents the Master OMD (M-OMD) algorithm for efficiently computing population-dependent Nash equilibria in MFGs, allowing us to address challenges such as diverse initial populations and common noise. Future work includes exploring convergence proofs and extensions to more complex models such as multi-population MFGs.

\section*{Acknowledgements}

M.L. is affiliated with the NYU Shanghai Center for Data Science and NYU-ECNU Institute of Mathematical Sciences at NYU Shanghai.

\bibliographystyle{ieeetr}
\bibliography{ref}

\newpage
\appendix
%

\url{https://drive.google.com/file/d/1nXKRwdVhSw-HogyzcnoGbz_9Pxw6wx3b/view?usp=sharing}
\newpage

\section{Additional algorithm framework}
See Algo. \ref{algorithm2} for classic fictitious play (FP) and Algo. \ref{algorithm3} for classic online mirror descent (OMD).

\begin{algorithm}[htbp]
\caption{Classic Fictitious Play (FP)}\label{algorithm2}
\SetAlgoLined
\SetKwInOut{Input}{Input}
\SetKwInOut{Output}{Output}
\Input{Number of iterations $K$, initial policy $\pi^0$}
\Output{Final average distribution $\bar{\mu}^K$ and policy $\bar{\pi}^K$}

\For{$k = 0, \ldots, K$}{
  \textbf{Forward Update:} Compute $\mu^k = \mu^{\pi^{k-1}}$\;
  
  \textbf{Average Distribution Update:} For every timestep $n$, update:
  
  $\quad \bar{\mu}_n^k(x) = \frac{1}{k}\sum_{i=1}^k \mu_n^i(x)$\;
  
  $\quad\quad= \frac{k-1}{k}\bar{\mu}^{k-1}_{n}(x) + \frac{1}{k}{\mu}_n^k(x)$\;
  
  \textbf{Best Response Computation:} Compute a BR $\pi^k$ against $\bar{\mu}^k$, e.g., by computing $Q^{*,\bar{\mu}^k}$ and then taking $\pi_n^k(. \mid x)$ as a distribution over $\arg\max Q^{*,\bar{\mu}^k}(x,\cdot)$ for every $n, x$\;
}

\end{algorithm}

\begin{algorithm}[htbp]
\caption{Classic Online Mirror Descent (OMD)}\label{algorithm3}
\SetAlgoLined
\SetKwData{Left}{left}\SetKwData{This}{this}\SetKwData{Up}{up}
\SetKwFunction{Union}{Union}\SetKwFunction{FindCompress}{FindCompress}
\SetKwInOut{Input}{input}\SetKwInOut{Output}{output}
\begin{flushleft}
\textbf{Input:} {learning rate parameter $\tau$; number of iterations $K$; timestep $n$; \\
Initialize Q table $\left(\bar{q}_n^0\right)_{n=0, \ldots, N_T}$, e.g. with $\bar{q}_n^0(x, a)=0$ for all $n, x, a$} 
\end{flushleft}
\begin{flushleft}
\textbf{Output:} {policy $\pi_K$ and final regularized ${\bar{q}^K}$} 
Initialize:$(\bar{q}_n^0){n=0,\ldots,N_T}$, e.g. with $\bar{q}_n^0(x, a)=0$, for all $n$, $x$, $a$.
\end{flushleft}
\begin{flushleft}
Let the projected policy be: $\pi_n^0(a \mid x)=\operatorname{softmax}(\bar{q}_n^0(x,\cdot))(a)$ for all $n$, $x$, $a$.

\For{$k=1, \ldots, K$}{ 

Forward Update: $\mu^k=\mu^{\pi^{k-1}}$.

Backward Update: $Q^k=Q^{\pi^{k-1}, \mu^k}$.

Update the regularized  $Q$:\;

\qquad $\bar{q}_n^k(x, a)=\bar{q}_n^{k-1}(x, a)+\frac{1}{\tau} Q_n^k(x, a)$\;

\qquad $\pi_n^k(a \mid x)=\operatorname{softmax}(\bar{q}_n^k(x,\cdot))(a)$\;

}
\Return{$(\bar{q}^K, \pi^K)$}
\end{flushleft}
\end{algorithm}

\section{More details on the Q-function updates}

\subsection{Equivalent formulation of the Q-function updates}
\label{proof:thm}
In the main text, we propose a Theorem~\ref{thm:equivalence} that is the key foundation of our algorithm as well as the corresponding proof. We denote by $\mathrm{KL}$ the Kullback-Leibler divergence: for $\pi_1,\pi_2 \in \Delta_{\mathcal{A}}$, $\mathrm{KL}\left(\pi_1 \| \pi_2\right)=\left\langle\pi_1, \ln \pi_1-\ln \pi_2\right\rangle$, which is also used in other regularized RL algorithms to make the training stage more stable, see e.g. \cite{vieillard2020munchausen,vieillard2020leverage,cui2021approximately,lauriere2022scalable}. By this theorem, we can give the $Q$ update and $\widetilde{Q}^k$ update equations respectively: 
 
\begin{equation}
\label{oldq}
\begin{aligned}
     {Q}_n^k(x, \mu_n^{k},\cdot)=&r_n(x, \mu_n^{k})+ \\
    &
    \gamma  \sum_{a'} 
    \pi_{n+1}^k (x, \mu_{n+1}^{k},a') Q^k_{n+1}(x, \mu_{n+1}^{k},a')
\end{aligned}
\end{equation}

\begin{equation}
\label{newq}
\begin{aligned}
     \tilde{Q}_n^k\left(x, \mu_n^k, a \right)&=r_n\left(x, \mu_n^k, a\right)  +\tau \ln \pi_n^{k-1}\left(x, \mu_n^k, a\right) \\
    &  +\gamma \sum_{a'} 
    \pi_{n+1}^k\left(x, \mu_{n+1}^k, a'\right) \Bigg[\tilde { Q }_{ n + 1}^{ k }(x, \mu_{n+1}^k, a')\\
    &\qquad\qquad - \tau \ln \pi_{n+1}^{k-1}\left(x, \mu_{n+1}^k, a'\right) \Bigg]
\end{aligned}
\end{equation}

The main idea behind Theorem~\ref{thm:equivalence} is that we establish the connection between the two equations below. We first prove the two equalities~\eqref{softmax1} and \eqref{softmax2}, then prove the equivalence between the right hand side of both equations. See Appx.~
\ref{proof:softmax}. 
\begin{equation}
\label{softmax1}
     \operatorname{softmax}\left(\frac{1}{\tau}\sum_{i=0}^k Q^i\right)=\underset{\pi}{\operatorname{argmax}}\left\langle\pi, Q^k\right\rangle-\tau \mathrm{KL}\left(\pi \| \pi^{k-1}\right)
\end{equation}
\begin{equation}
\label{softmax2}
    \operatorname{softmax}\left(\frac{1}{\tau} \tilde{Q}^k\right) = \underset{\pi}{\operatorname{argmax}}\left\langle\pi, \tilde{Q}^k\right\rangle -\tau\langle\pi , \ln \pi\rangle 
\end{equation}

Here we use the $\mathrm{softmax}$ instead of solving $\mathrm{argmax}$ in~\eqref{softmax2}, because solving  $\mathrm{argmax}$ is not always guaranteed when using Deep RL, and the $\mathrm{softmax}$ implicitly contains the greedy step which can be exactly computed, which also benefits the convergence \cite{vieillard2020leverage}.

Therefore, for the cost function of Q update in Algo. \ref{algo: algorithm1}: 

\begin{equation}
    \mathbb{E}\left|\tilde{Q}^k_\theta\left(\left(n, x_n,\mu^k_n\right), a_n\right)-T\right|^2
\end{equation}
where $T$ is defined in \eqref{eq:target-Tn}.

\subsection{Proof of Theorem~\ref{thm:equivalence}}
\label{proof:softmax}

\begin{proof}

In order to prove the result,  we will first expand the expressions for $\operatorname{softmax}\left(\frac{1}{\tau} \sum_i^k Q^i\right)$ and $\operatorname{softmax}\left(\frac{1}{\tau} \widetilde{Q}^k\right)$ to obtain~\eqref{softmax1} and~\eqref{softmax2}. Then, we will show that the right-hand sides of~\eqref{softmax1} and~\eqref{softmax2} are equivalent.

\noindent{\bf Step 1.} We prove the expansion of~\eqref{softmax1}, i.e. 
\begin{equation}
\begin{aligned}
    \operatorname{softmax}&\left(\frac{1}{\tau} \sum_{i=0}^k Q^i(x,\mu^k_n,n)\right)
    \\
    &=\underset{\pi}{\operatorname{argmax}}\left\langle\pi, Q^k(x,\mu^k_n,n)\right\rangle- \\
    &\qquad\tau \mathrm{KL}\left(\pi(\cdot \mid x,\mu^k_n,n) \| \pi^{k-1}(\cdot \mid x,\mu^k_n,n)\right)
\end{aligned}
\label{eq16}
\end{equation}

Here $Q^k$ is the standard Q-function at iteration $k$, as defined in~\eqref{oldq}. $\pi^{k-1}$ is the policy learned in iteration $k-1$.  $\mu^{k}$ is the mean field induced by the policy $\pi^{k-1}$.

First, we define a new function, denoted as $F$, which corresponds to the right-hand side of \eqref{eq16}. For brevity, we exclude the explicit inputs of $\pi$ and $Q$, since the optimization process solely focuses on optimizing the parameter $\pi$. Hence, we express this simplification as $Q = Q(x,\mu_n,n)$. Since the policy is the probability distribution, an additional constraint is needed to guarantee that the sum of $\pi$ is 1.  

\begin{equation}
\begin{gathered}
\underset{\pi}{\operatorname{argmax}} \, F(\pi)=\underset{\pi}{\operatorname{argmax}}\left\langle\pi, Q^k\right\rangle-\tau \mathrm{KL}\left(\pi \| \pi^{k-1}\right) \\
\text { s.t. } \quad \boldsymbol{1}^{\top} \pi=1,
\end{gathered}
\end{equation}
where $\boldsymbol{1}$ denotes a vector full of ones, of dimension the number of actions.

We then introduce the Lagrange multiplier $\lambda$ and the Lagrangian $L$, defined as:
\begin{equation}
    L(\pi,\lambda)=\left\langle\pi, Q^k\right\rangle-\tau \mathrm{KL}\left(\pi \| \pi^{k-1}\right)+\lambda\left(\boldsymbol{1}^{\top} \pi-1\right)
\label{eq:lagrange}
\end{equation}

Now, by finding the equilibrium, we need to find a saddle point of~\eqref{eq:lagrange}. So let us compute the partial derivatives of $L$. Proceeding formally, we obtain:
\begin{equation}
    \begin{aligned}
\frac{\partial L(\pi, \lambda)}{\partial \pi} & =Q^k-\tau \frac{\partial\langle\pi \ln \pi\rangle}{\partial \pi}+\tau \frac{\left\langle\pi, \ln \pi^{k-1}\right\rangle}{\partial \pi}+\lambda(\mathbf{1}) \\
& =Q^k-\tau\left(\ln \pi+1-\ln \pi^{k-1}\right)+\lambda \mathbf{1} \\
\frac{\partial L(\pi, \lambda)}{\partial \lambda} & =\mathbf{1}^{\top} \pi-1
\end{aligned}
\label{eq:second_pd}
\end{equation}

Note that, for every $\lambda$, the function $\pi \mapsto L(\pi,\lambda)$ is concave, as the sum of a linear function and the negative of the KL divergence (and since the KL divergence is convex). 

Taking $\frac{\partial L(\pi,\lambda)}{\partial \pi}=0$, we find that the optimum satisfies:
\begin{equation}
\begin{aligned}
\pi & =e^{\frac{1}{\tau} (Q^k +\lambda\boldsymbol{1}) -1} \cdot \pi^{k-1} \\
& =e^{\frac{1}{\tau} (Q^k +\lambda\boldsymbol{1})-1} \cdot e^{\frac{1}{\tau} \cdot (Q^{k-1} +\lambda\boldsymbol{1})-1} \ldots \\
& =e^{\frac{1}{\tau}\left(Q^k+Q^{k-1}+\cdots+Q^0+(k+1)\lambda\boldsymbol{1}\right)-(k+1)} \\
& =e^{\frac{1}{\tau} \sum_{i=0}^k Q^i} \cdot e^{-(k+1)} \cdot e^{\frac{(k+1)}{\tau} \lambda\boldsymbol{1}} \\
& =e^{\frac{1}{\tau} \sum_{i=0}^k Q^i}\cdot {C_1} \cdot {C_2{(\lambda})}
\end{aligned}
\label{eq:first_pd2}
\end{equation}
where $C_1$ is a constant which equals to $e^{-(k+1)}$, $C_2$ is the function of the Lagrange multiplier $\lambda$. In order to satisfy the constraint  $\frac{\partial L(\pi,\lambda)}{\partial \lambda}=0$, i.e. $\operatorname{sum}(\pi)=\sum \pi=1$, note that $\lambda$ is a scalar, and $\frac{1}{\tau} \sum_{i=0}^k Q^i$ is a vector with dim of $|A|$, if \eqref{eq:lambda} holds, then the optimization problem \eqref{prob:optimization1} is solved. 

\begin{equation}
    C_1 \cdot C_2(\lambda) = \frac{1}{\sum_{\frac{1}{\tau} \sum_{i=0}^k Q^i} e^{\frac{1}{\tau} \sum_{i=0}^k Q^i}}
\label{eq:lambda}
\end{equation}

Thus, $\pi$ is a softmax function as follows, 
\begin{equation}
\begin{aligned}
\pi= \operatorname{softmax}\left(\frac{1}{\tau} \sum_{i=0}^k Q^i\right)
\end{aligned}
\end{equation}

\noindent{\bf Step 2. } Define a new function $G(\pi)$ 

\begin{equation}
    \underset{\pi}{\operatorname{argmax}} G(\pi) = \underset{\pi}{\operatorname{argmax}}\left\langle\pi, \tilde{Q}^k\right)-\tau\langle\pi , \ln \pi\rangle 
\end{equation}
following the same way as solving \eqref{prob:optimization1}, we can prove the second equality as needed. i.e the expansion of \eqref{softmax2}:  \
\begin{equation}
    \underset{\pi}{\operatorname{argmax}}\left\langle\pi, \tilde{Q}^k\right\rangle-\tau\langle\pi , \ln \pi\rangle =  \operatorname{softmax}\left(\frac{1}{\tau} \tilde{Q}^k\right) 
\end{equation}

\noindent{\bf Step 3. }

We now prove the equivalence between the two right hand sides. 
Let $\tilde{Q}^k=Q^k+\tau \ln \pi_{k-1}$.
Then 
\begin{equation}
    \begin{aligned}
&\underset{\pi}{\operatorname{argmax}}\left\langle\pi \cdot Q^k\right\rangle-\tau \mathrm{KL}\left\langle\pi \| \pi^{k-1}\right\rangle 
\\
& =\underset{\pi}{\operatorname{argmax}}\left\langle\pi, \tilde{Q}^k-\tau \ln \pi_{k-1}\right\rangle-\tau \mathrm{KL}\left(\pi \| \pi^{k-1}\right) \\
& =\underset{\pi}{\operatorname{argmax}}\left\langle\pi, \tilde{Q}^k\right\rangle-\tau\langle\pi, \ln \pi\rangle
\end{aligned}
\end{equation}
Therefore, Theorem~\ref{thm:equivalence} is proved.

\end{proof}

\section{More details about algorithms}

\subsection{Env2: Beach bar}
\paragraph{Example 2: Beach bar without Common Noise }
The Beach bar environment, introduced in~\cite{perrin2020fictitious} represents agents moving on a beach towards a bar. The goal for each agent is to (as much as possible) avoid the crowd but get close to the bar. The dynamics are the same as in the exploration examples. Here we consider that the bar is located at the center of the beach, and that it is not possible to go beyond the domain. The reward function is:
$
r\left(x, a, \mu\right)={d_{bar}}\left(x\right)-\frac{\left|a\right|}{|\mathcal{X}|}-\log \left(\mu\left(x\right)\right), 
$
where $d_{bar}$ indicates the distance to the bar, the second term penalizes movement so the agent moves only if it is necessary, and the third term penalizes the fact of being in a crowded region. Here, we consider $\mathcal{X}$ with $11$ (1D) or $11\times11$ (2D) states. 

\begin{figure} [htb]
\centering
\subfloat[Evolution process in 2D]{
\begin{minipage}[t]{0.9\linewidth}
\centering
\includegraphics[width=3in]{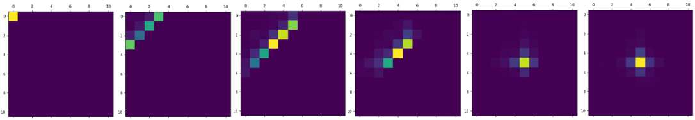}
\end{minipage}
}%
\newline
\subfloat[Evolution process in 1D]{
\begin{minipage}[t]{0.5\linewidth}
\centering
\includegraphics[width=1.6in]{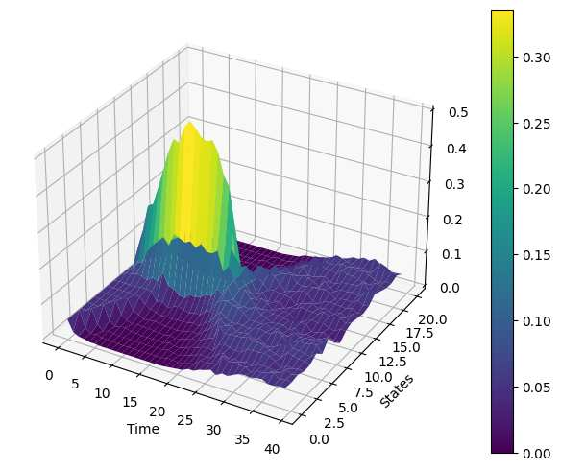}
\end{minipage}
}%
\subfloat[Exploitability (fixed $\mu_0$)]{
\begin{minipage}[t]{0.5\linewidth}
\centering
\includegraphics[width=1.45in]{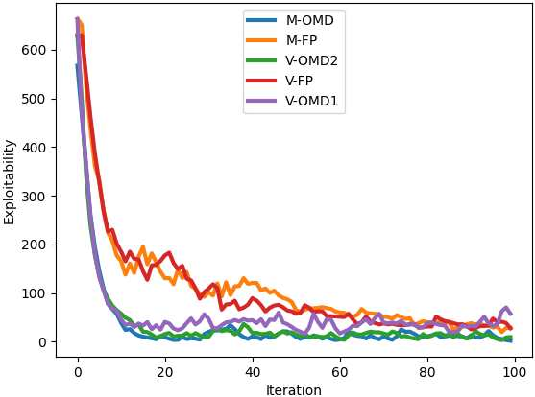}
\end{minipage}
}\
\subfloat[Exploitability (multiple $\mu_0$)]{
\begin{minipage}[t]{0.55\linewidth}
\centering
\includegraphics[width=1.45in]{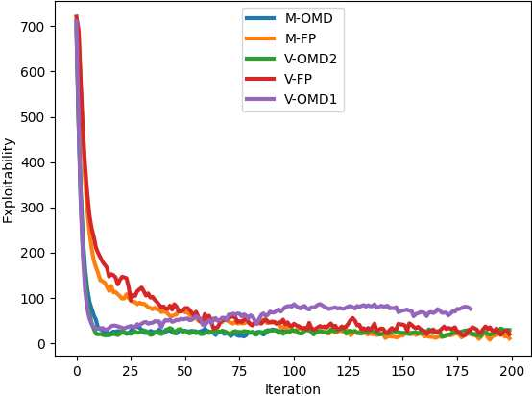}
\end{minipage}
}%
\subfloat[Exploitability (multiple $\mu_0$)]{
\begin{minipage}[t]{0.45\linewidth}
\centering
\includegraphics[width=1.55in]{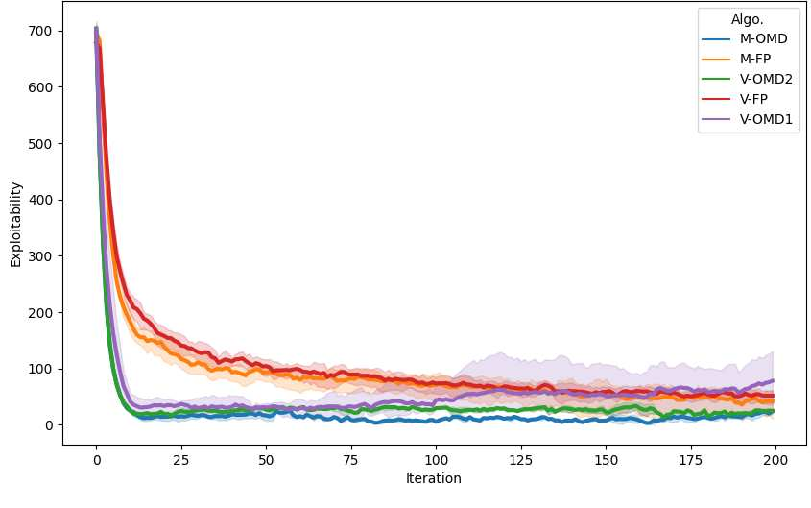}
\end{minipage}
}
\caption{Example 2 in Env2: Beach bar problem. (a) and (b) show the distribution evolution for 2D and 1D case. (c), (d) and (e): exploitability for 2D case with: (c) when training with fixed $\mu_0$, (d) when training with different $\mu_0$, and (e) when training with different $\mu_0$ and averaging over 5 runs.
}
\label{fig:beach_bar}
\end{figure}

The results are shown in Fig.~\ref{fig:beach_bar}. The agents are always attracted towards the bar, whose location is independent of the population distribution. So the impact of starting from a fixed distribution or from various distributions is expected to be relatively minimal, which is also what we observe numerically. (a) is for a 2D domain and we see that the distribution spreads while the agents move towards the bar to avoid large crowds, and then focuses on the bar's location. (b) is for a 1D model but we add an extra difficulty: the bar closes at time $n=20$ and hence the population goes back to a uniform distribution (due to crowd aversion). This shows that the policy is aware of time. (c) and (d) show the performance when training over multiple $\mu_0$. Here again, M-OMD performs best.

\subsection{Env3: Linear-Quadratic}
\label{app:lq}
\paragraph{Example 2: Linear-Quadratic without Common Noise}
\label{sec:lq}
In this section, we provide results on a LQ model without common noise. It is a linear quadratic (LQ) model, which is a classical setting that has been studied extensively; see e.g. \cite{carmona2013control}, \cite{bensoussan2016linear}. A discretized version was introduced in~\cite{perrin2020fictitious}. It is a 1D model, in which the dynamics are: $x_{n+1}=x_n+a_n \Delta_n+\sigma \epsilon_n \sqrt{\Delta_n}$, where $\mathcal{A}=\{-M, \ldots, M\}$, corresponding to moving by a corresponding number of states (left or right). The state space is $\mathcal{X}=\{-L, \ldots, L\}$, of dimension $|\mathcal{X}|$ =$2L-1$. To add stochasticity into this model, $\epsilon_n$ is an additional noise will perturb the action choice with $\epsilon_n \sim \mathcal{N}(0,1)$, but was discretized over $\{-3 \sigma, \ldots, 3 \sigma\}$. The reward function is:
\[
    r_n\left(x,a,\mu\right)=\big[-\tfrac{1}{2}\left|a\right|^2+q a\left(m-x\right)-\tfrac{\kappa}{2}\left(m-x\right)^2\big] \Delta_n
\]
where $m=\sum_{x \in \mathcal{X}} x \mu(x)$ is the first moment of population distribution which serves as the reward to encourage agents to move to the population's average but also tries to keep dynamic movement. The terminal reward is $r_{N_T}\left(x, a, \mu\right)=-\frac{c_{\text {term }}}{2}\left(m-x\right)^2$. Here we used $\sigma=1$, $N_T=30$, $\Delta_n=1$, $q=0.01$, $\kappa=0.5$, $K=1$ $M=3$, $L=20$, $c_{term}=1$.

\begin{figure}[htb]
\centering
\subfloat[Evolution process]{
\includegraphics[width=0.3\linewidth]{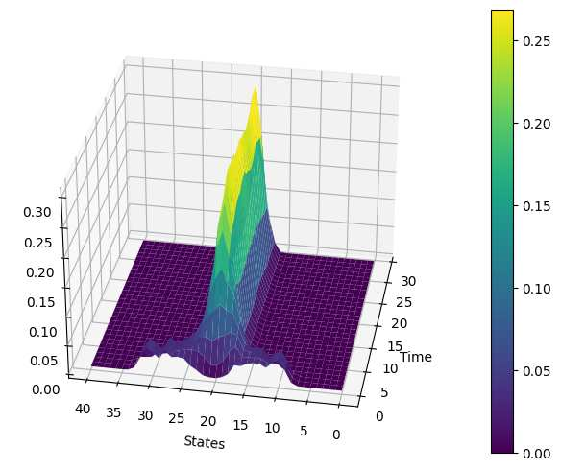}
}%
\subfloat[Exploitability (fixed $\mu_0$)]{
\includegraphics[width=0.25\linewidth]{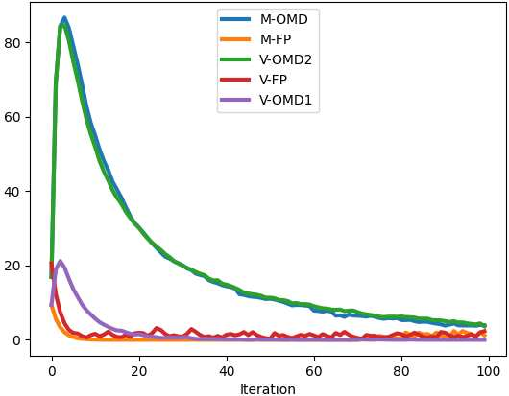}
}\,\,%
\subfloat[Exploitability (multiple $\mu_0$)]{
\includegraphics[width=0.29\linewidth]{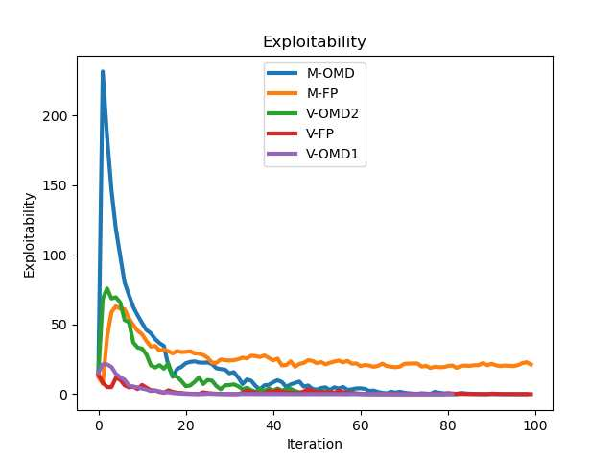}
}%
\caption{Example 2 in Env3: Linear quadratic model. (a) shows the evolution of the population using the policy learned by the M-OMD algorithm, starting from two Gaussian distribution pairs, and then accumulating into the center of the population. (b) and (c) shows the averaged exploitability obtained during training over one fixed initial distribution and five initial distributions, respectively.} 
\label{fig:lq}
\end{figure}

Compared with the tasks mentioned above, solving this LQ game is considerably easier and convergence occurs with only a few iterations. However, this gives rise to some counter-intuitive phenomena in the numerical results. Our algorithm rely on modified Bellman equations, which incorporate an additional term as a regularizer to prevent a rapid change of policy during training. Consequently, our algorithm for LQ does not converge to the Nash equilibrium as fast as FP. This desirable degeneration is shown in Fig.~\ref{fig:lq}, where FP and V-OMD1 demonstrate faster convergence compared to V-OMD2 and M-OMD. In the case of V-OMD1, the parameters we used are small regularized coefficients than M-OMD and V-OMD2 (see sweeping results in Fig.~\ref{fig: sweep} in Appx.) resulting in faster convergence than our algorithms, though still slower than FP. Regarding the exploitability in multiple initial distribution training, in theory, the population should gravitate towards the center of the whole population. However, the results indicate that even vanilla FP or OMDs can decrease to zero. Our analysis is that since the moving cost is cheaper than the rewards agents receive, vanilla policies can learn a strategy that moves all agents to a specific position regardless of the initial distributions. Our tests also revealed that if the cost of moving is too high, all algorithms learn a policy that keeps agents stationary. 

This explanation can also be revealed in LQ with common noise case \ref{fig:lq_cn1}. In that case, all OMD and FP algorithms show the same convergence rate. It is because the common noise serves as an anchor trajectory, the population distribution evolution

 The convergence of FP and OMD In \ref{fig:lq_cn1} keeps the same scale of convergence while FP owns faster convergences than OMD in \ref{fig:lq}. It is due to the dimensions scale of the spaces. In the small dimensional LQ, NE policy is instantly more like a deterministic behavior where the regularizer in OMD prohibits the policy changes faster. In the large-dimensional LQ problem, the early stage of population

Finding more appropriate values for the LQ model's parameters to demonstrate the influence of population-dependence is deferred to future research.

\subsection{Buffer Size Sweeping}
\begin{figure}[htbp!]
  \centering
    \begin{minipage}[t]{0.8\linewidth}
    \centering
    \includegraphics[width=2.5in]{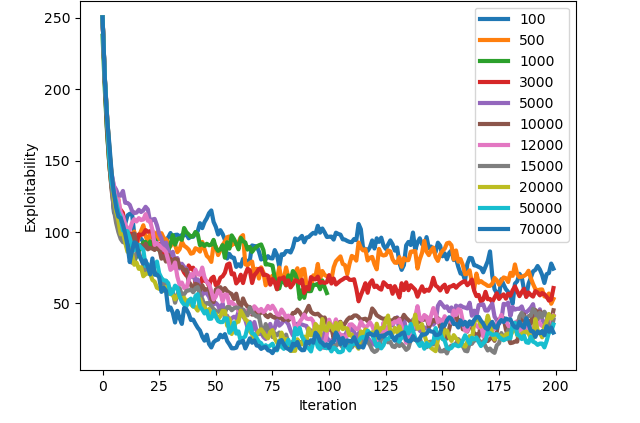}
    \end{minipage}%
  \caption{Exploitability vs iteration number for various buffer sizes, using our M-OMD algorithm, in exploration of four connected room task. Small sizes lead to the forgetting of some $\mu_0$ and hence poor performance (see Step 2 in Algo.~\ref{algo: algorithm1}). } 
\label{fig:buffer_replay}
\end{figure}

See Fig. \ref{fig:buffer_replay}.

\subsection{Ad-hoc teaming}
\label{sec:adhoc-teaming}
\begin{figure}[htbp]
  \centering
  \subfloat[Small team joining]{
    \includegraphics[width=0.45\columnwidth]{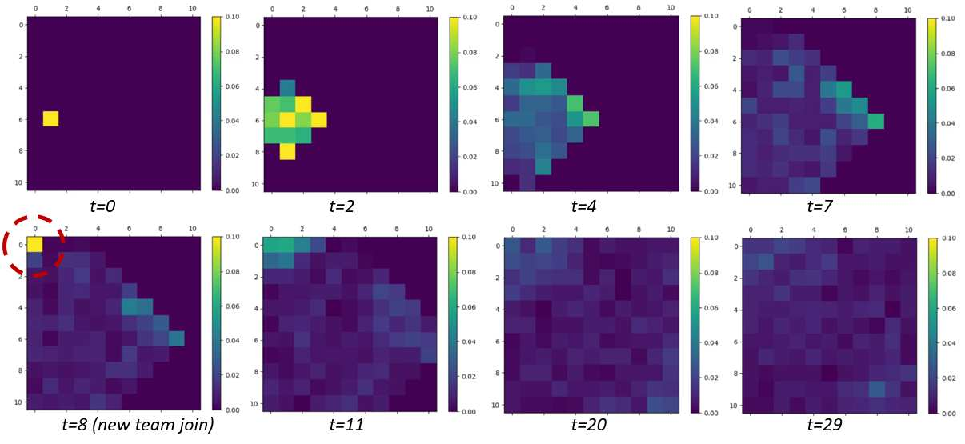}
  }
  \subfloat[Large team joining]{
    \includegraphics[width=0.45\columnwidth]{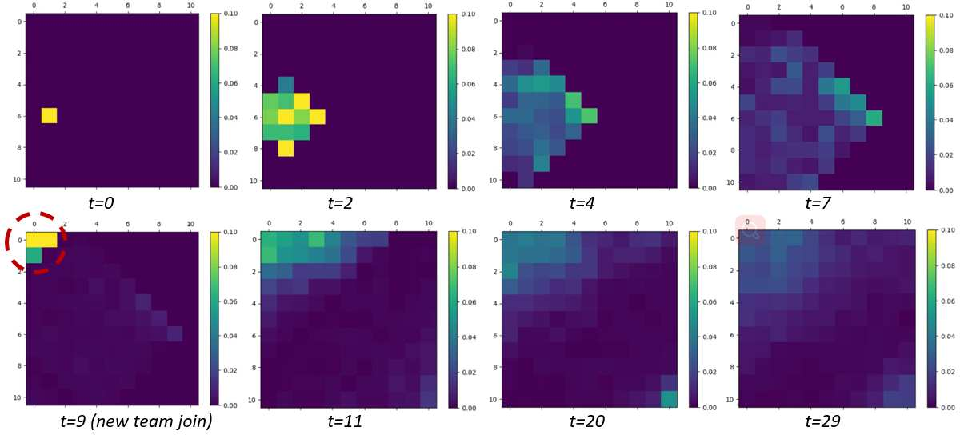}
  }
  \caption{Ad-hoc teaming test simulates new team joining into the current team (500 agents) during the evolutive process of the population. The simulation contains two different teams, one is a small team joining (200 agents), another is a large group team joining (2500 agents)}
  \label{fig:ad-hoc}
\end{figure}

In this paper, we introduce a novel testing case for the master policy, referred to as Ad-hoc teaming, inspired by ad-hoc networks in telecommunication. Ad-hoc teaming simulates the scenario where additional agent groups join the existing team during execution, resembling spontaneous and temporary formations without centralized control or predefined network topology. By incorporating the concepts of distribution and time awareness, the policy should enable multiple teams to join at any timestep, ultimately achieving the Nash equilibrium. Fig. \ref{fig:ad-hoc} illustrates two instances where different agent groups join the current team during execution. One group consists of a small number of agents, causing minimal impact on the overall population distribution, while the other group comprises a larger number of agents, significantly altering the distribution. As shown in Fig.~\ref{fig:ad-hoc}, the population still leads to uniform distribution after the small team joins. However, when a large group joins the current team, the final distribution at the terminal timestep deviates from the expected uniform distribution. The reasons can be attributed to two aspects. Firstly, the time left on the horizon is not sufficient to allow ad-hoc agents to spread out. Secondly, the generalization learning limits. During the training process, the population initially starts from a single area and subsequently spreads across the map. Consequently, the policy networks' distribution awareness is implicitly limited to the spread-out tendency. However, the random emergence of ad-hoc teams disrupts this spread-out distribution of the population. To address this issue, it is necessary to incorporate additional scenarios like ad-hoc teaming during training. This paper presents an initial exploration of such a testing scenario in real life, leaving a more comprehensive investigation for future research.

\begin{figure}[h!]      
  \centering
  \includegraphics[width=0.4\textwidth]{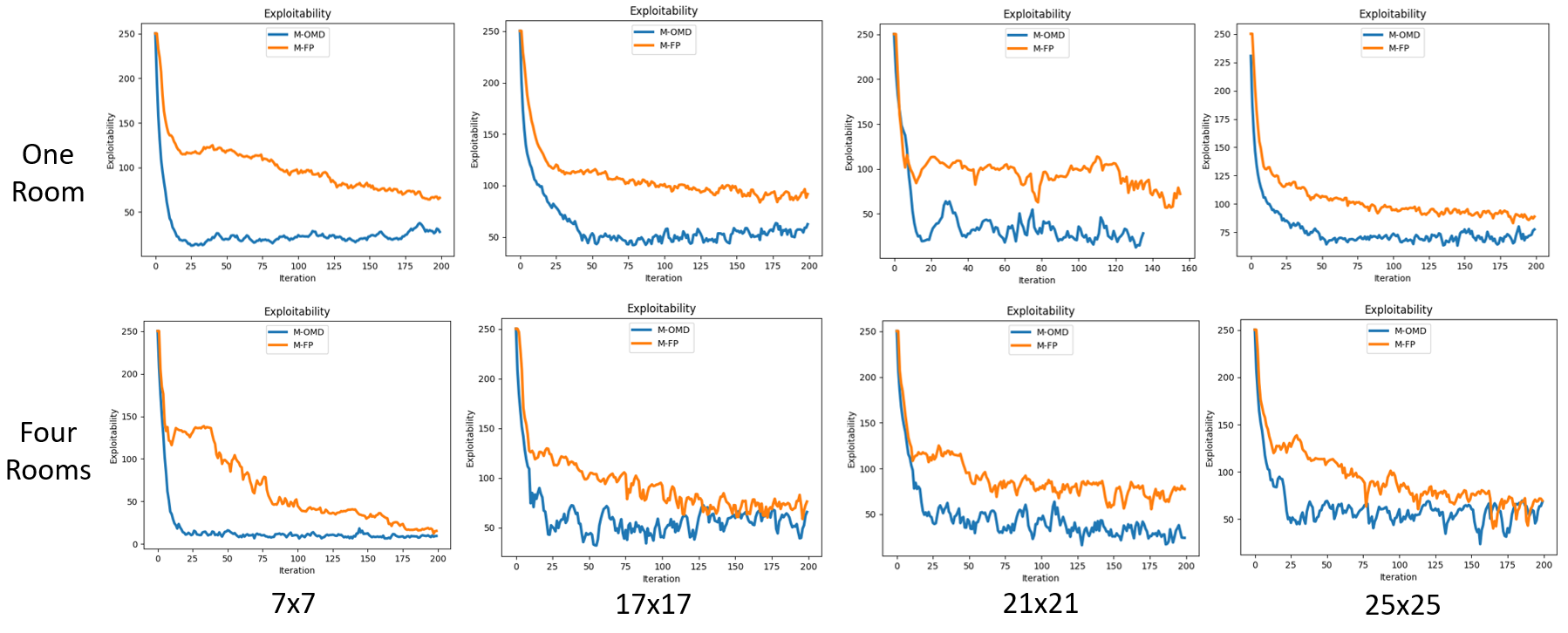}
  \caption{Exploitability versus map size. We compared two master policies: M-OMD (ours) and M-FP(SOTA) in five different map dimensions. Due to the time horizon of games, a map size larger than 25x25 is not meaningful as agents cannot explore even half the map before termination.}
  \label{fig:map_size}
\end{figure}

\begin{figure}[htbp]
  \centering
  \subfloat[Memory vs Map size]{
    \includegraphics[width=0.4\columnwidth]{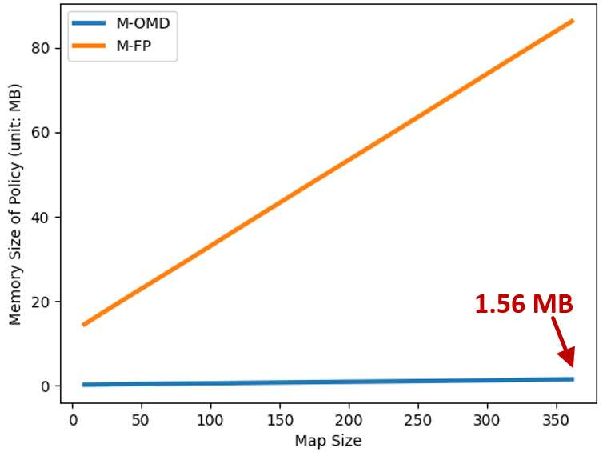}
  }
  \subfloat[Memory vs Iteration training]{
    \includegraphics[width=0.4\columnwidth]{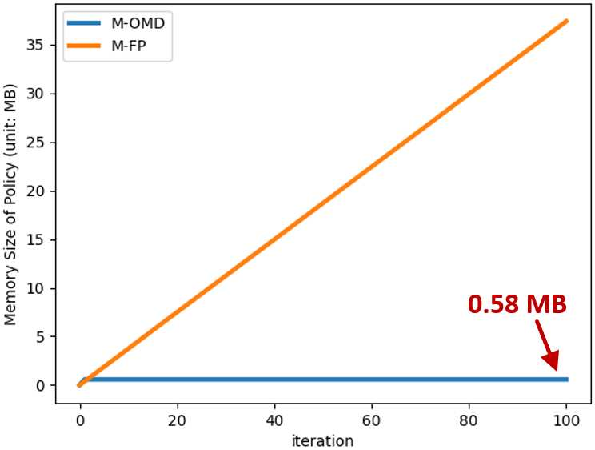}
  }
  \caption{Model size comparison for M-OMD (ours) and M-FP.}
  \label{memory_cost}
\end{figure}

\subsection{Training set and testing set}
\begin{figure}[htbp]
  \centering
  \subfloat[Training set]{
    \includegraphics[width=0.5\textwidth]{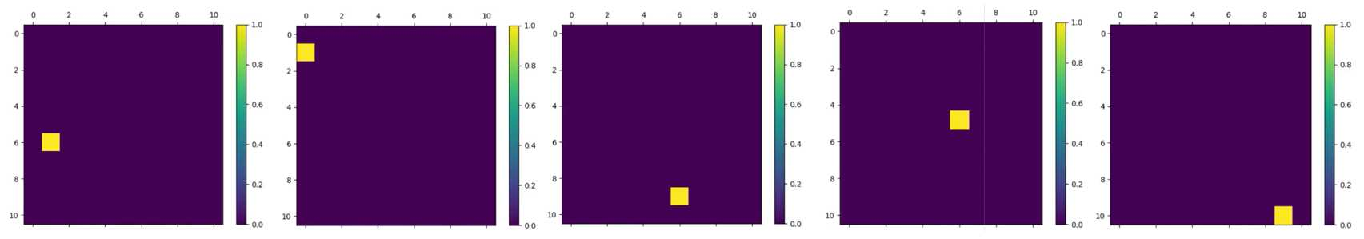}
  }
  \newline
  \subfloat[Testing set]{
    \includegraphics[width=0.5\textwidth]{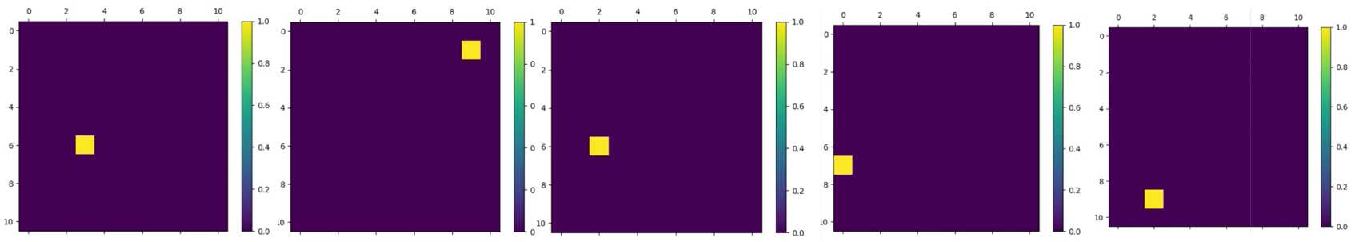}
  }
  \caption{Training and testing sets for Beach bar task \& Exploration in One room task}
  \label{fig:trainset_uniform}
\end{figure}

\begin{figure}[htbp]
  \centering
  \subfloat[Training set]{
    \includegraphics[width=0.5\textwidth]{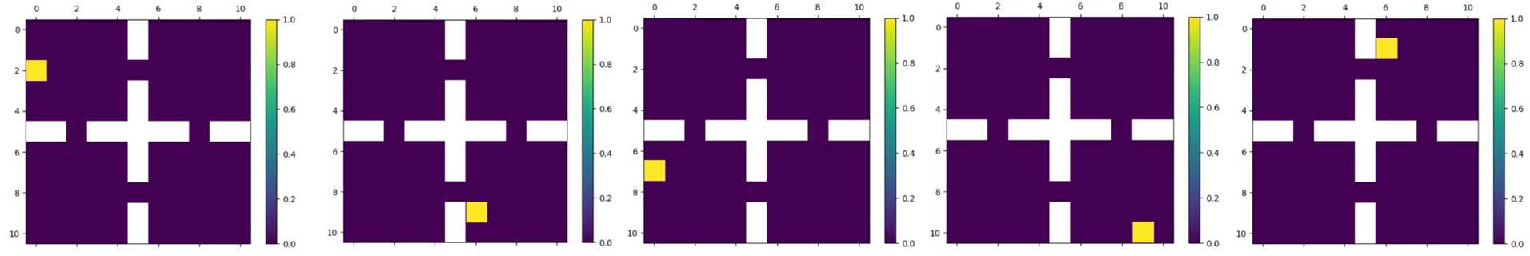}
  }
  \newline
  \subfloat[Testing set]{
    \includegraphics[width=0.5\textwidth]{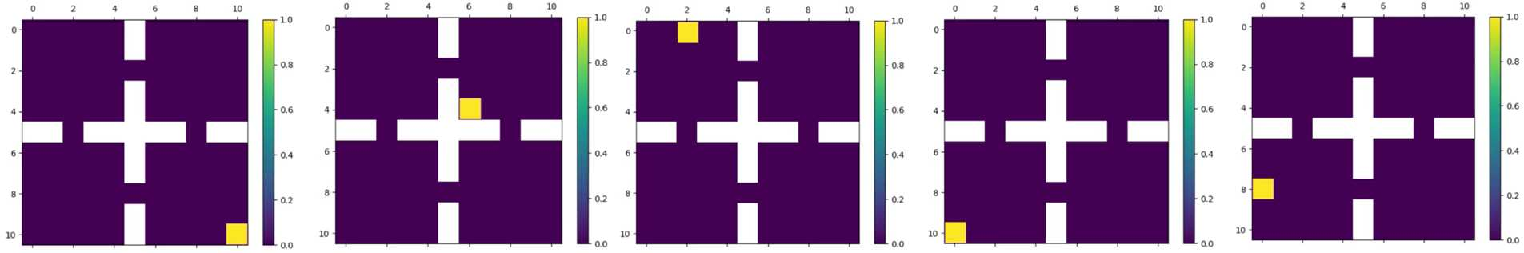}
  }
  \caption{Training and testing sets for Exploration in Four Rooms task}
  \label{fig:trainset_exploration2}
\end{figure}

To validate the effectiveness of the learned master policy, we adopt the approach described in \cite{Perrin2022General} to construct separate training and testing sets. This section presents the five training sets used to learn the policy and five testing sets utilized to evaluate its performance. The distributions for the Beach bar task and Exploration in one room task are depicted in Fig. \ref{fig:trainset_uniform}, while Fig. \ref{fig:trainset_exploration2} illustrates the distributions for the Exploration in four rooms task. We provide a summary of the exploitability of the testing sets in Table ~\ref{table:exploitability_testing}. It shows that all algorithms exhibit higher exploitability than the training set, as evidenced by the exploitability curves in the training figures at the final iteration. We attribute this to overfitting and insufficient training data, which are classic challenges in the field of machine learning. The M-OMD results demonstrate a significant reduction in exploitability during training, although it does not maintain the same level during testing. Insufficient data implies a limited representation of diverse initial distributions, causing the neural network to struggle with changes in population distribution, which is also revealed in the Ad-hoc teaming tests to some extent. However, it is important to note that overfitting and insufficient amounts of training data are common issues in ML, which do not undermine the feasibility of our algorithm. Addressing these challenges and improving the effectiveness of training are the topics for future research. The reason why vanilla policies perform better than master policies during testing has been discussed in the LQ section.

\subsection{Hyperparameters for experiments}
See table \ref{table:hyperparam} for the training parameters for all five algorithms.
\begin{table}[htbp]
\caption{Hyperparameters used for training Exploration task with map size:11x11}
\label{table:hyperparam}
\begin{center}
\begin{small}
\begin{tabular}{p{1.5 cm}ccccc}
\toprule
Algorithm & M-OMD & M-FP & V-FP & V-OMD1 & V-OMD2 \\
\midrule
NN Arch & mlp & mlp & mlp & mlp & mlp \\
Neurons per Layer & 64*64 & 64*64 & 64*64 & 64*64 & 64*64 \\
Horizon & 30 & 30 & 30 & 30 & 30 \\
Agents num & 500 & 500 & 500 & 500 & 500 \\
Max Steps per iteration & 30000 & 30000 & 30000 & 30000 & 30000 \\
OMD $\tau$ & 50 & N/A & N/A & 5.0 & 50\\
OMD $\alpha$ & N/A & N/A & N/A & 1.0 & 1.0\\
Freq to update target & 4 & 4 & 4 & 4 & 4 \\
Exploration Fraction & 0.1 & 0.1 & 0.1 & 0.1 & 0.1 \\
$\gamma$ & 0.99 & 0.99 & 0.99 & 0.99 & 0.99 \\
Batch Size & 32 & 32 & 32 & 32 & 32 \\
Gradient Steps & 1 & 1 & 1 & 1 & 1 \\
\bottomrule
\end{tabular}
\end{small}
\end{center}
\end{table}

\subsection{Hyperparameter sweeping}
\begin{figure}[htbp]
  \centering
  \subfloat[M-OMD]{
    \includegraphics[width=0.45\columnwidth]{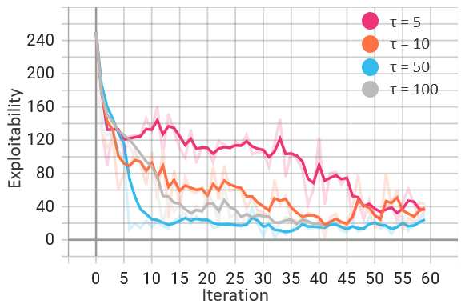}
  }
  \hfill 
  \subfloat[V-OMD1]{
    \includegraphics[width=0.45\columnwidth]{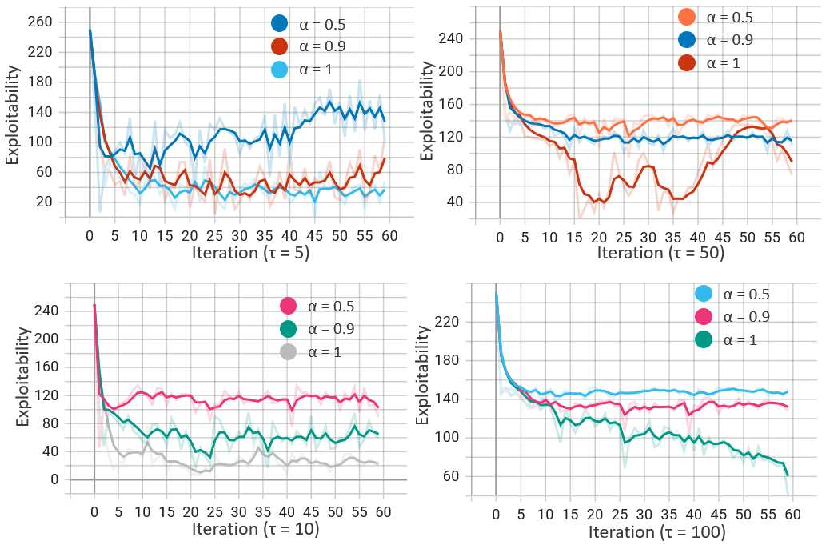}
  }
  \caption{Hyperparameter sweeping}
  \label{fig: sweep}
\end{figure}

We provide the sweeping curves of hyperparameter $\tau$, both ours and V-OMD1, and $\alpha$, only in V-OMD1 \cite{lauriere2022scalable} in this section. See Fig. \ref{fig: sweep}.

\subsection{Computational time}
\label{sec:comput-time}
\begin{figure}[htb]
\centering
\subfloat[Total time per iteration]{
\begin{minipage}[t]{0.45\linewidth}
\centering
\includegraphics[width=1.5in]{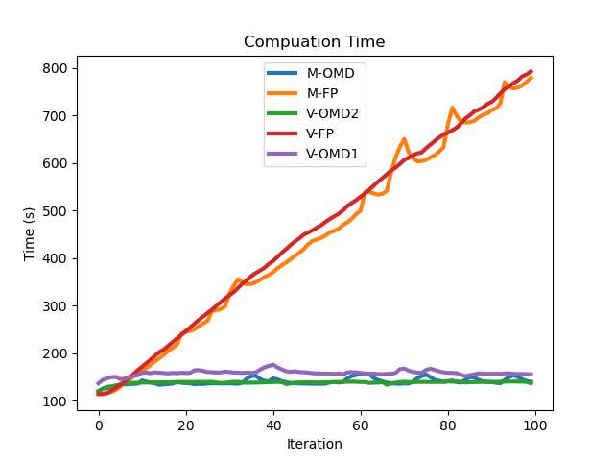}
\end{minipage}%
}%
\subfloat[Learn policy]{
\begin{minipage}[t]{0.45\linewidth}
\centering
\includegraphics[width=1.5in]{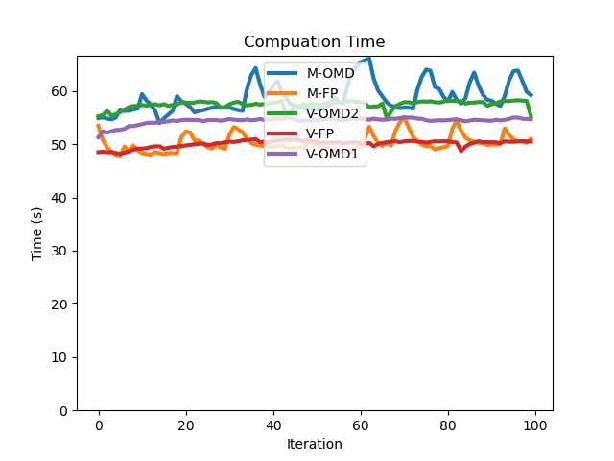}
\end{minipage}%
}%
\newline
\subfloat[Update distribution]{
\begin{minipage}[t]{0.45\linewidth}
\centering
\includegraphics[width=1.5in]{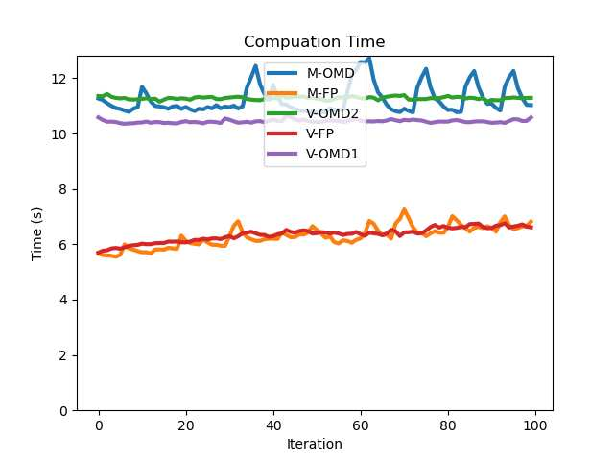}
\end{minipage}%
}%
\subfloat[Compute exploitability]{
\begin{minipage}[t]{0.45\linewidth}
\centering
\includegraphics[width=1.5in]{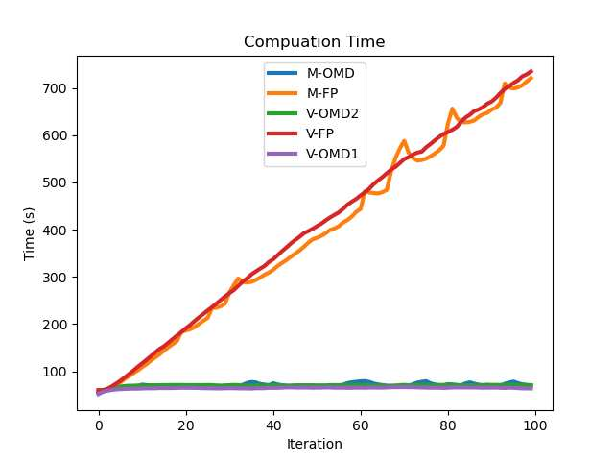}
\end{minipage}
}%
\caption{Computational time comparison (Exploration task). Figures correspond to three steps in each iteration, (b)learning the best response in FP or evaluating Q in OMD; (c) updating population distribution based on new learned policy;(d) calculating Exploitability}
\label{fig:time_computation}
\end{figure}

To calculate the exploitability during training, FP needs to use all history policies to execute while our algorithm only uses a single policy network, which results in a linear increase in computation cost for FP during training but not for our algorithm. Therefore, even though the policy learning time per iteration of our algorithm would be slightly longer than FP-based algorithms, it still saves much more computational time considering the convergence speed and computation of exploitability. See Fig.\ref{fig:time_computation} for details.

\subsection{Training with 30 initial distributions}
\begin{figure}[htbp]
  \centering
{
    \includegraphics[width=0.5\columnwidth]{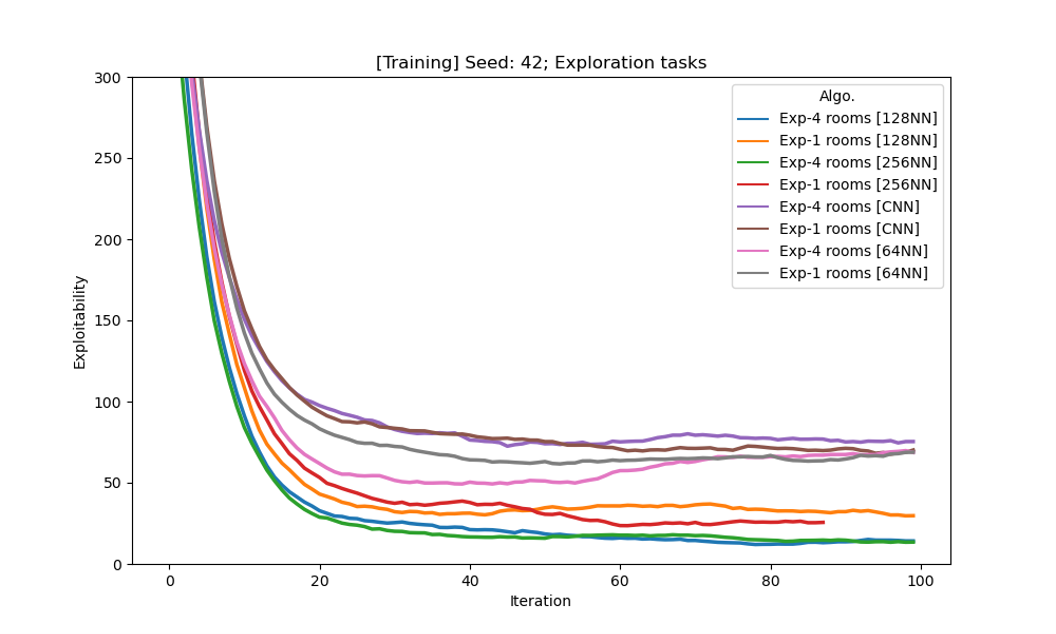}
  }
  \caption{Training on two exploration tasks with 30 distributions. Four different architectures are tested for Master OMD algorithm}
  \label{fig:exploitability_nn_archtecture}
\end{figure}

\begin{figure}[htb]
\centering
\subfloat[Exploration in 1 room (30 $\mu_0$) ]{
\includegraphics[width=0.45\linewidth]{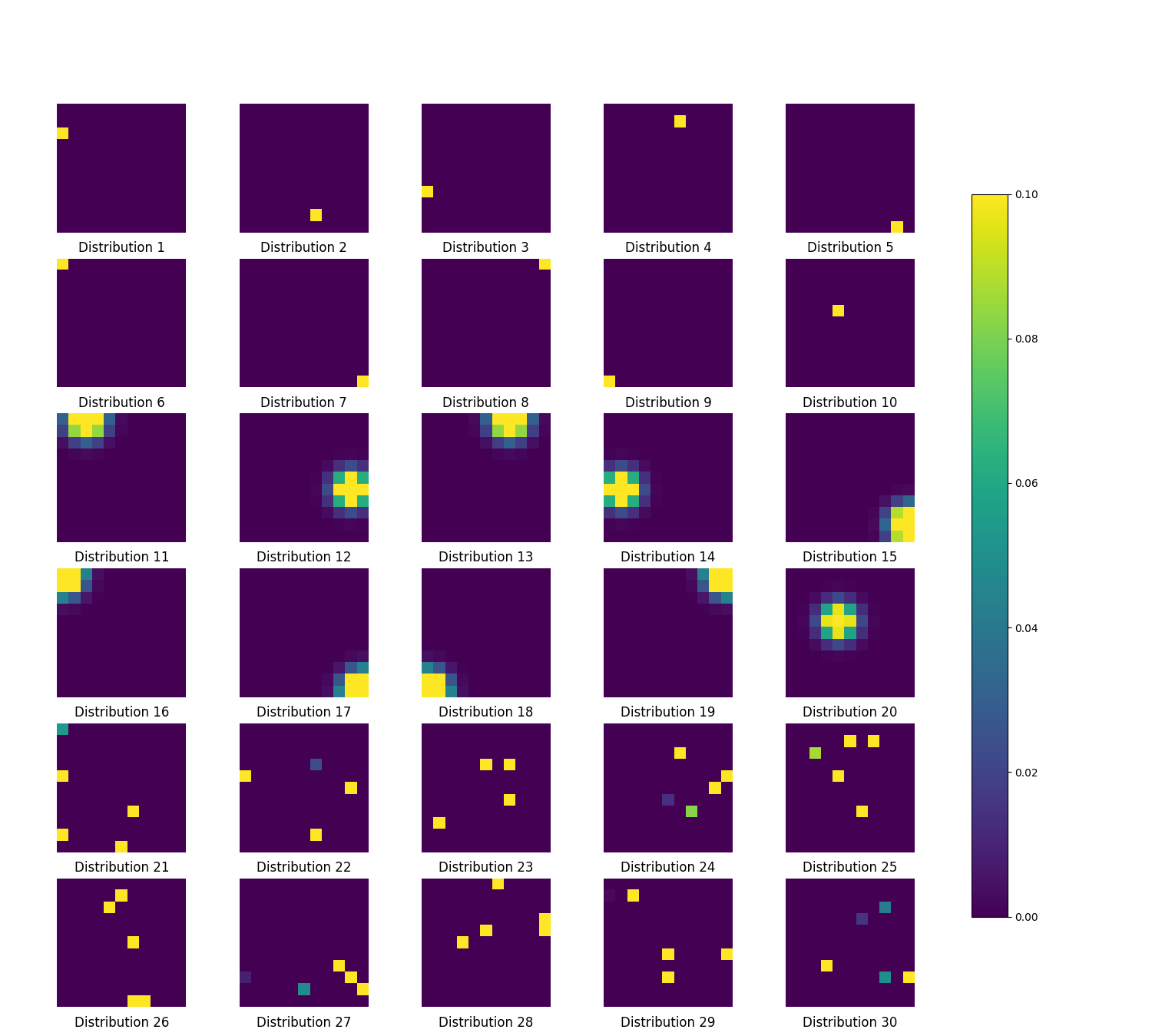}
}%
\subfloat[Exploration in 4 rooms (30 $\mu_0$) ]{
\includegraphics[width=0.45\linewidth]{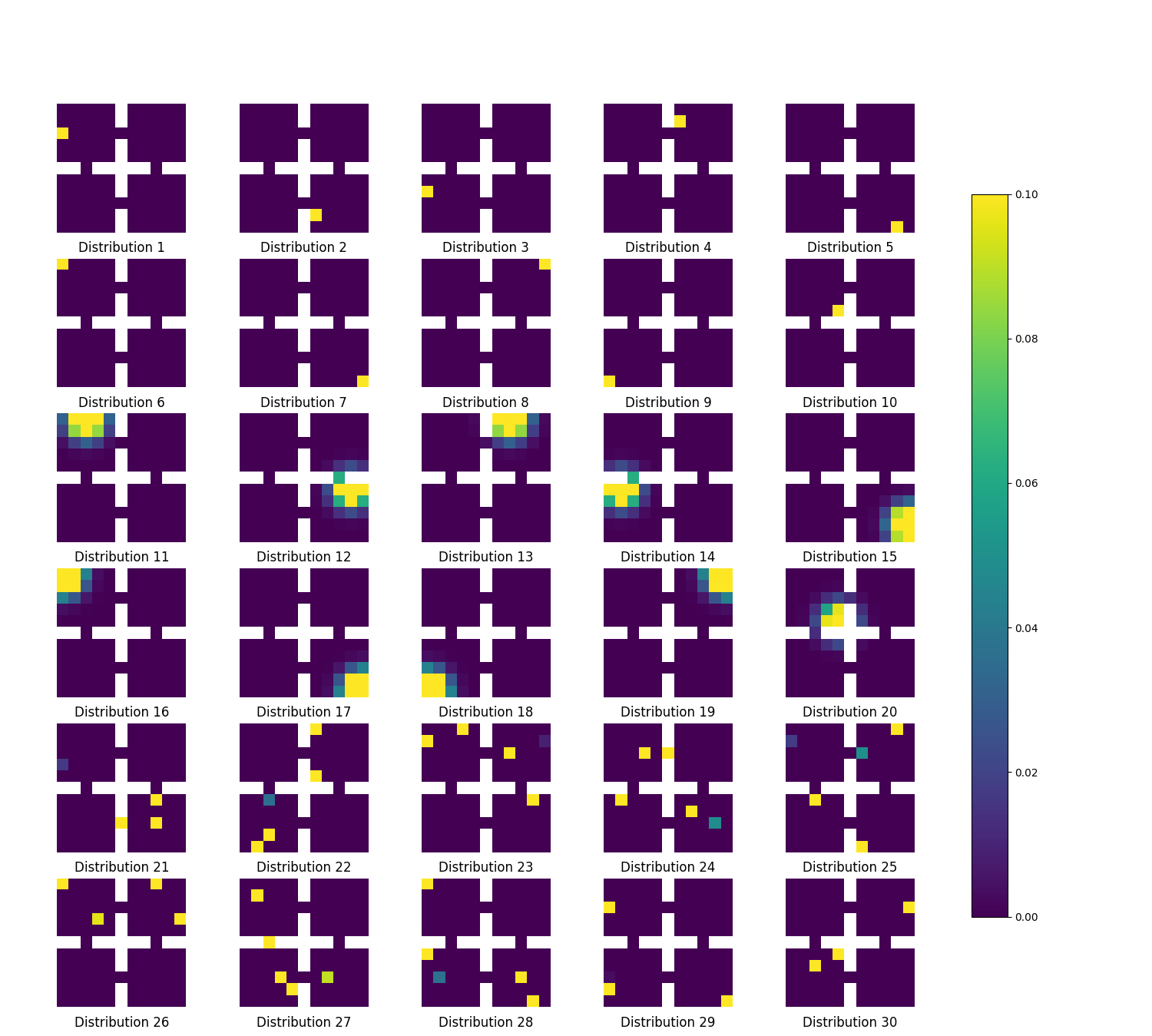}
}%

\caption{Training sets with 30 distributions for two exploration tasks (a) Exploration in One room (b) Exploration in Four room
} 
\label{fig:training_set_dis30}
\end{figure}

In the main text, we demonstrate that our algorithm proficiently handles five distributions, as illustrated in Fig \ref{fig:trainset_uniform}. To extend our exploration of its adaptability across a broader spectrum of distributions, we examine an ensemble of 30 distributions depicted in Fig \ref{fig:training_set_dis30}. This set comprises 10 distributions originating from fixed points, 10 following Gaussian distributions, and 10 distributed across random points. To assess the impact of policy architecture on performance, we evaluated four distinct architectures: three MLP-based architectures with configurations of $64\times64$, $128\times128$, and $256\times256$ layers, respectively, alongside a CNN-based architecture featuring two convolutional layers ($32\times64$, with kernel sizes of 5 and 3) followed by a fully connected layer. As evidenced in Fig \ref{fig:exploitability_nn_archtecture}, the $64\times64$ MLP architecture struggles to converge when handling 30 distributions. Consequently, we adopt the $256\times256$ architecture as our benchmark for further investigation. 

\begin{figure}[htb]
\centering
\subfloat[Training stage: Exploration in One room (30 $\mu_0$) ]{
\includegraphics[width=0.45\linewidth]{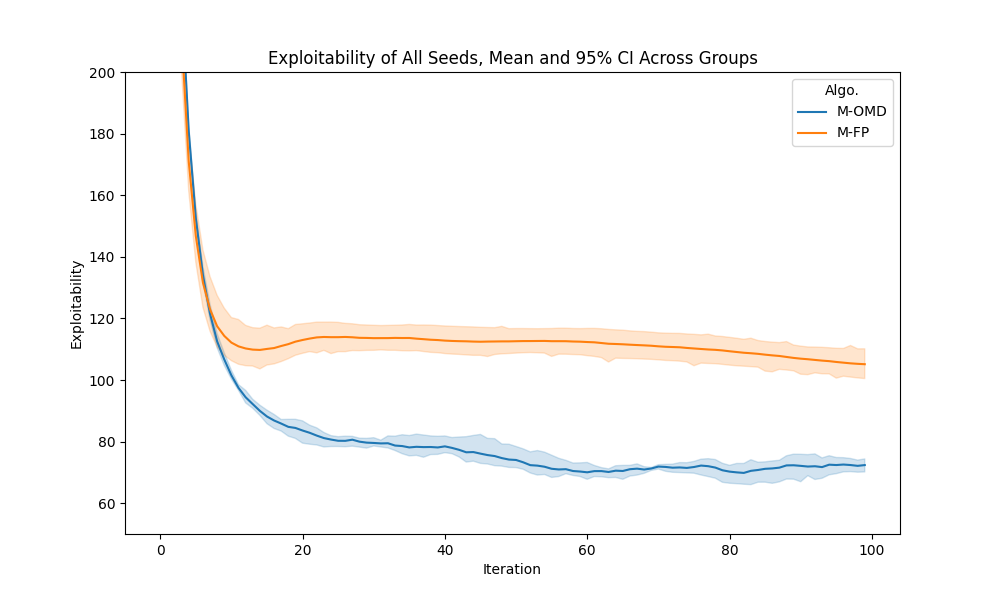}
}%
\subfloat[Testing stage: Exploration in Four rooms (30 $\mu_0$)]{
\includegraphics[width=0.45\linewidth]{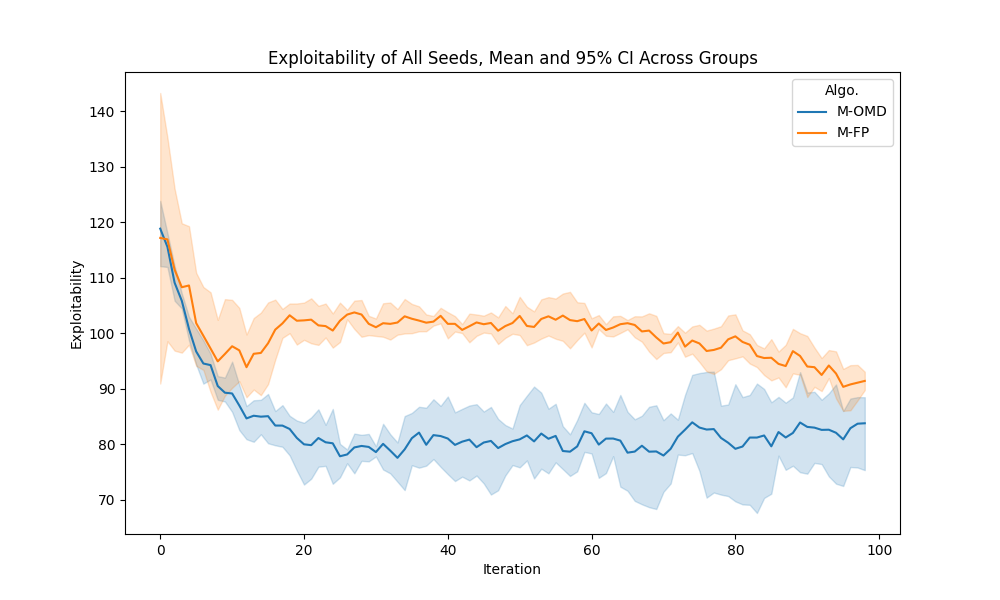}
}%

\caption{Exploration in One Room task: Exploitability for 30 initial distributions with the best response based on dynamic programming with MLP architecture $[256\times256]$. (a) is the comparison between Master FP and Master OMD (ours) in the training stage and (b) is the comparison between Master FP and Master OMD (ours) in the testing stage
} 
\label{fig:exploitability_dis30_1room}
\end{figure}

\begin{figure}[htb]
\centering
\subfloat[Training stage: Exploration in Four rooms (30 $\mu_0$) ]{
\includegraphics[width=0.45\linewidth]{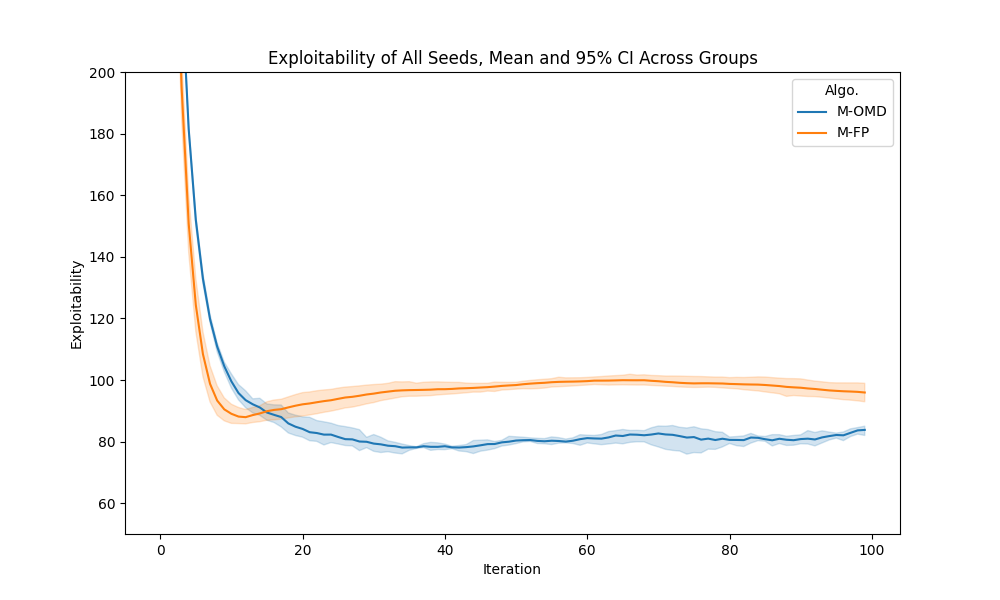}
}%

\subfloat[Testing stage: Exploration in Four rooms (30 $\mu_0$) ]{
\includegraphics[width=0.45\linewidth]{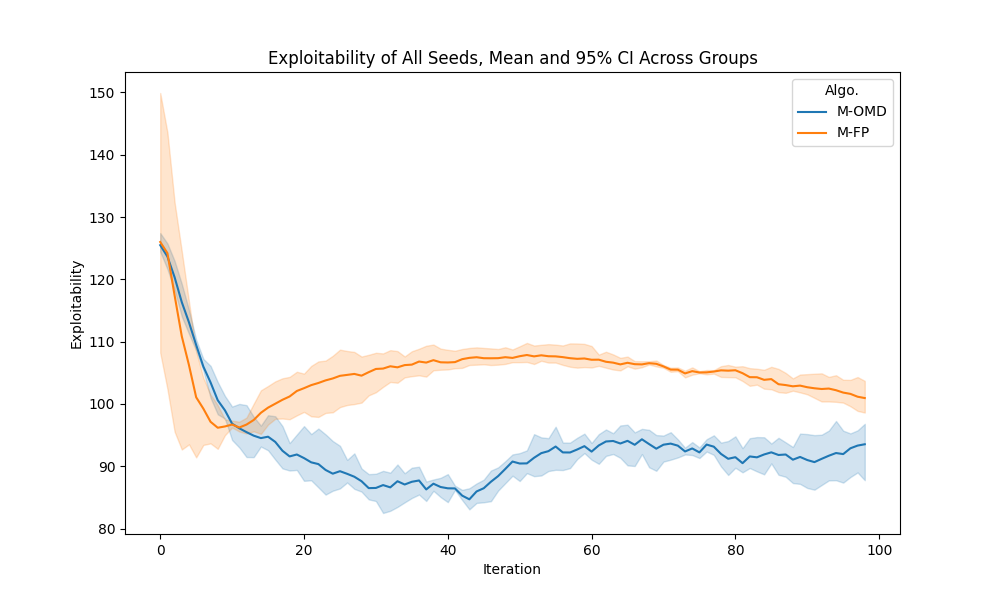}
}%

\caption{Exploration in Four Rooms task: Exploitability for 30 initial distributions with the best response based on dynamic programming with MLP architecture $[256\times256]$. (a) is the comparison between Master FP and Master OMD (ours) in the training stage and (b) is the comparison between Master FP and Master OMD (ours) in the testing stage
} 
\label{fig:exploitability_dis30_4rooms}
\end{figure}

While the utilization of DQN for learning the optimal response is prevalent in Deep Mean Field Games (MFG), we aim to distinguish clearly between the DQN-derived best response and the authentic best response. To achieve this, we employ dynamic programming to solve best response, enabling us to precisely compute the true exploitability. We conducted experiments using both the Master Fictitious Play (M-FP) and Master Online Mirror Descent (M-OMD) algorithms, maintaining identical network architectures across two exploration tasks, each tested with five seeds: 42, 3407, 303, 109, and 312. As shown in Fig \ref{fig:exploitability_dis30_1room} and Fig \ref{fig:exploitability_dis30_4rooms}, our approach outperforms the population-based FP algorithm in terms of true exploitability, but also demonstrates advantages in computational efficiency, execution time, and model compactness, as previously highlighted.

\begin{figure}[htb]
\centering
\subfloat[Exploration in 1 room (30 $\mu_0$) ]{
\includegraphics[width=0.45\linewidth]{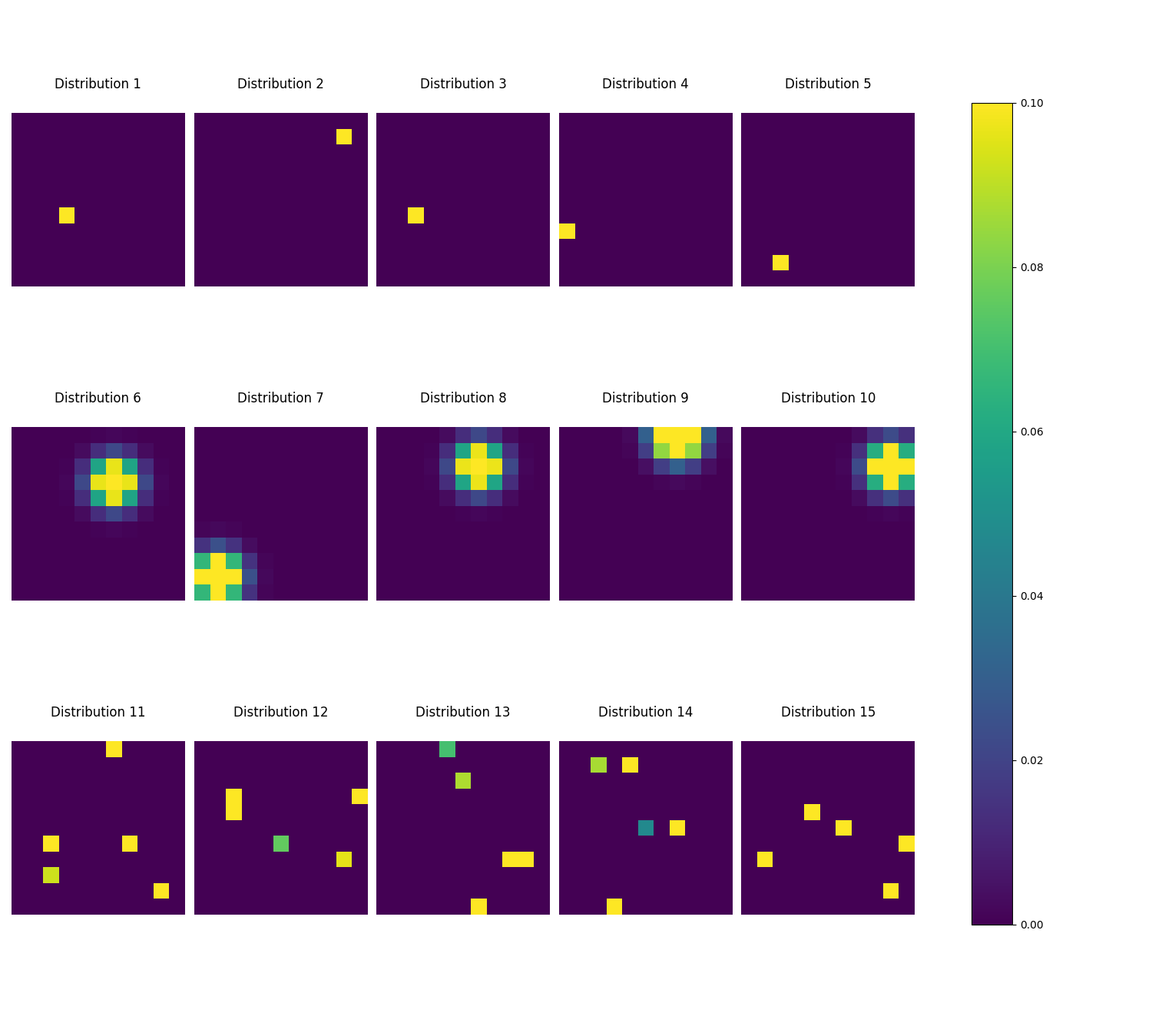}
}%
\subfloat[Exploration in 4 rooms (30 $\mu_0$) ]{
\includegraphics[width=0.45\linewidth]{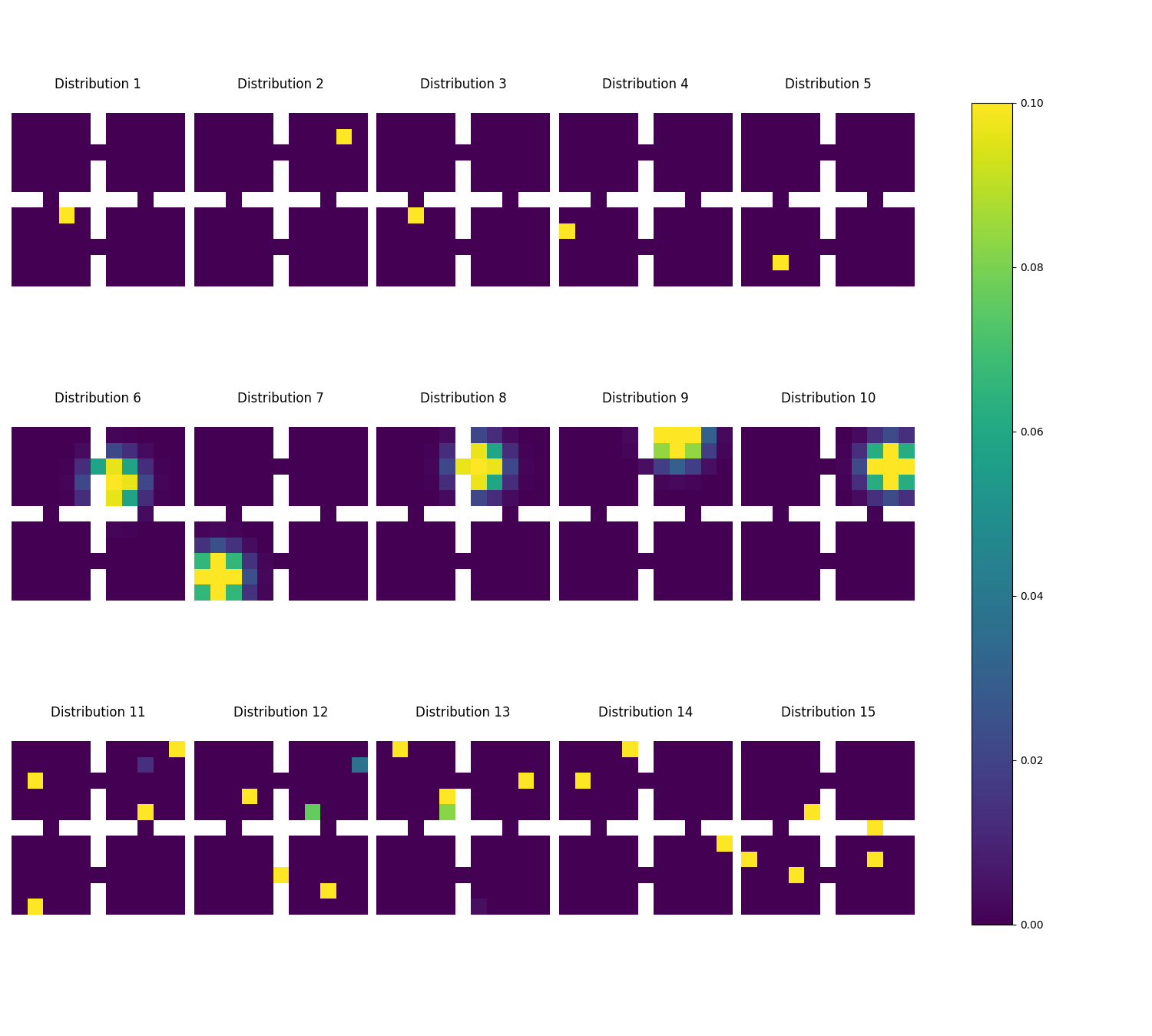}
}%

\caption{Testing sets with 30 distributions for two exploration tasks (a) Exploration in One room (b) Exploration in Four room
} 
\label{fig:tesing_set_dis30}
\end{figure}

\end{document}